\documentclass[10pt,journal]{IEEEtran}
\usepackage{color, soul}
\usepackage{graphicx}
\usepackage{microtype}
\usepackage{verbatim}
\usepackage{epstopdf}
\usepackage{setspace}
\usepackage[ruled,vlined,linesnumbered]{algorithm2e}
\usepackage{amsmath}
\usepackage{amssymb}
\usepackage{amsthm}
\usepackage{mathrsfs}
\usepackage{extarrows}
\usepackage{bbm}
\usepackage{array}
\usepackage{booktabs}
\usepackage{flushend}
\usepackage{cite}
\usepackage{multirow}
\usepackage[caption=false,font=footnotesize]{subfig}
\PassOptionsToPackage{hyphens}{url}
\usepackage{xcolor}
\usepackage[
    colorlinks=true,
    linkcolor=blue,
    citecolor=blue,
    urlcolor=blue,
    breaklinks=true
]{hyperref}
\newcommand{\bicaption}[4][]{\caption{#4}\if\relax\detokenize{#1}\relax\else\label{#1}\fi}
\newcommand{\subfigure}[2][]{\subfloat[#1]{#2}}

\let\origincludegraphics\includegraphics
\renewcommand{\includegraphics}[2][]{%
	\IfFileExists{#2}{\origincludegraphics[#1]{#2}}{%
		\fbox{\parbox[c][3cm][c]{0.9\linewidth}{\centering Missing figure:\\ \texttt{\detokenize{#2}}}}%
	}%
}

\begin{document}
	\title{\textsc{SCORP}: Scene-Consistent Multi-agent Diffusion Planning with Stable Online Reinforcement Post-Training for Cooperative Driving}
	
	\author{Haojie Bai, Aimin Li, \IEEEmembership{Member, IEEE}, Ruoyu Yao, Xiongwei Zhao, Tingting Zhang, \IEEEmembership{Member, IEEE}, Xing Zhang, Caixiong Li, Lin Gao, \IEEEmembership{Senior Member, IEEE}, and Jun Ma, \IEEEmembership{Senior Member, IEEE}%
		\IEEEcompsocitemizethanks{%
            \IEEEcompsocthanksitem The project page is available at: \url{https://zebai9.github.io/SCORP/}.
			\IEEEcompsocthanksitem Haojie Bai, Xiongwei Zhao, Tingting Zhang, and Lin Gao are with the School of Information Science and Technology, Harbin Institute of Technology (Shenzhen), Shenzhen 518071, China (e-mail: hjbai@stu.hit.edu.cn, gaol@hit.edu.cn).
			\IEEEcompsocthanksitem Aimin Li is with the Middle East Technology University (METU), Ankara, 06800, Turkiye (e-mail: aimin@metu.edu.tr). Aimin Li contributes equally to this work.
			\IEEEcompsocthanksitem Ruoyu Yao and Jun Ma are with the Robotics and Autonomous Systems Thrust, Hong Kong University of Science and Technology (Guangzhou), Guangzhou 511453, China (e-mail: ryao092@connect.hkust-gz.edu.cn, jun.ma@ust.hk).
			\IEEEcompsocthanksitem Xing Zhang and Caixiong Li are with the School of Computer Science and Technology, Qinghai University, Xining, 810016, China.
		}%
	}
	\maketitle
	
		\begin{abstract}

            Cooperative driving is a safety- and efficiency-critical task that requires the coordination of diverse, interaction-realistic multi-agent trajectories. Although existing diffusion-based methods can capture multimodal behaviors from demonstrations, they often exhibit weak scene consistency and poor alignment with closed-loop cooperative objectives. This makes post-training necessary for further improvement, yet achieving stable online post-training in reactive multi-agent environments remains challenging.
            In this paper, we propose \textsc{SCORP}, a scene-consistent multi-agent diffusion planner with stable online reinforcement learning (RL) post-training for cooperative driving. 
            For pre-training, we develop a scene-conditioned multi-agent denoising architecture that couples inter-agent self-attention with a dual-path conditioning mechanism: cross-attention provides direct scene-information injection, while AdaLN-Zero enables additional flexible and stable conditional modulation, thereby improving the scene consistency and road adherence of joint trajectories. For post-training, we formulate a two-layer Markov decision process (MDP) that explicitly integrates the reverse denoising chain with policy--environment interaction. We further co-design dense, well-shaped planning rewards and variance-gated group-relative policy optimization (VG-GRPO) to mitigate advantage collapse and gradient instability during closed-loop training. Extensive experiments show that SCORP outperforms strong open-source baselines on WOMD, with 10.47\%--28.26\% and 1.70\%--7.22\% improvements in core safety and efficiency metrics, respectively. Moreover, compared with alternative post-training methods, \textsc{SCORP} delivers significant and consistent gains in both driving safety and traffic efficiency, highlighting stable and sustained advances in closed-loop cooperative driving.

            

		\end{abstract}
	
	\section{Introduction}
	
	\subsection{Background and Motivation}
	

        Advanced \emph{Connected Autonomous Vehicles} (CAVs) have catalyzed the emergence of cooperative multi-agent driving paradigms, opening new possibilities for addressing key transportation challenges and potentially improving both safety and traffic efficiency~\cite{luo2023real,bai2025robust1}. By enabling information sharing and coordinated decision-making among vehicles, infrastructure, and cloud platforms, cooperative driving extends the capability of isolated single-vehicle autonomy and supports system-level intelligence in complex traffic environments\cite{11262787}. As such, cooperative driving is inherently a safety- and efficiency-critical task, especially in dense and interactive scenarios where the behaviors of multiple agents are coupled~\cite{bai2025robust2}.
        Central to this paradigm, \emph{multi-agent behavior modeling} underpins high-level cooperative decision-making by characterizing the interactive behaviors of multiple traffic participants. However, real traffic behavior is inherently stochastic and multimodal: even within the same scene, multiple interaction patterns can be reasonable and effective. This inherent uncertainty and diversity make it difficult to produce cooperative plans that are simultaneously realistic, safe, and efficient.
        Therefore, generating \emph{well-coordinated cooperative behaviors} while preserving multimodality and realistic interactions remains an open challenge, which in turn limits the deployment potential of cooperative autonomous driving systems~\cite{nayakanti2022wayformer}.

        Recent advances in diffusion models have introduced a powerful probabilistic paradigm for modeling complex multimodal distributions over driving behaviors and trajectories~\cite{janner2022planning,zheng2025diffusion}. By leveraging a forward diffusion and reverse denoising process and learning from large-scale human demonstrations via imitation learning, diffusion-based planners can capture dexterous behaviors and joint distributions over interactive multi-agent trajectories.
        However, existing methods often struggle to balance \emph{multi-agent interaction modeling} with \emph{scene-conditioned modeling}, leading to joint trajectories that may violate scene constraints~\cite{huang2024versatile,huang2025mdg}.
		Furthermore, these methods suffer from \emph{objective misalignment} and \emph{distribution shift}~\cite{lu2023imitation}. Behavior cloning (BC) mainly fits the data distribution and lacks the ability to explicitly enforce human preferences, expectations, or constraints, making it difficult to directly optimize safety and efficiency objectives in multi-agent systems. For example, safety-critical events such as collisions are exceedingly rare in human driving datasets, resulting in sparse supervision for learning safe interactions. Consequently, failures are more likely to occur under closed-loop execution or out-of-distribution scenarios. 
        Collectively, these issues hinder diffusion-based multi-agent planners from achieving safe and efficient closed-loop performance and limit their robustness and reliability in real-world scenarios~\cite{li2026plannerrft}.

		Reinforcement learning (RL) offers a promising avenue for further improving pretrained planning models by coupling sampling-based exploration with reward-driven policy optimization~\cite{gao2025rad,huang2025gen}. Recent studies suggest that RL post-training can enhance closed-loop planning performance and shape driving styles, thereby potentially promoting cooperative behaviors beyond what pretraining alone can achieve~\cite{peng2024improving,li2025finetuning,li2025recogdrive}. 
        In this paradigm, a pretrained planner serves as the actor, sampling diverse candidate futures that are scored by a reward function and iteratively refined through RL algorithms~\cite{guo2025deepseek,yu2025dapo}.
		However, most existing work focuses on \emph{offline} post-training, which is closer in spirit to reward-augmented supervised fine-tuning~\cite{zhang2025survey}. Without online interaction with reactive environments, constrained rollouts and limited exploration often yield only modest performance improvements~\cite{gao2025rad,li2025recogdrive}.
        
		More broadly, achieving satisfactory gains with RL post-training primarily depends on (i) online interaction, (ii) well-shaped rewards, and (iii) robust policy optimization.
		First, online interaction is naturally aligned with closed-loop execution, allowing the policy to explore interaction scenarios beyond the pretraining distribution and to correct behaviors under reward guidance, thereby mitigating performance degradation from distribution shift. 
		Second, well-shaped rewards provide fine-grained optimization signals for safety and efficiency-critical objectives while remaining compatible with realistic, human-like driving.
        Third, robust policy optimization further improves tolerance to reward noise and gradient variance, enabling stable and sustained performance gains.
        Nevertheless, realizing these benefits from online RL in fully closed-loop settings remains challenging due to non-stationary interactions, compounding rollout errors, and high-variance gradient estimates, which together impose stringent requirements on the stability and controllability of the post-training pipeline.
		
		\subsection{Solution and Contributions}
		To address these limitations, we propose \textsc{SCORP}, a scene-consistent multi-agent diffusion planner with \emph{online RL post-training}, tailored for closed-loop cooperative driving. 
        SCORP couples condition-enhanced diffusion pre-training with stable online RL post-training to continually improve closed-loop planning performance and cooperative behaviors.
        
		During pre-training, we build a multi-agent denoising network on the Diffusion Transformer~\cite{peebles2023scalable} to capture the scene-conditioned joint distribution of multi-agent trajectories while balancing inter-agent interaction modeling and scene conditioning.
		Specifically, we introduce \emph{AdaLN-Zero} adaptive modulation in conjunction with cross-attention. Through scene-driven feature modulation with zero-initialization, the model strengthens scene consistency and road adherence of multi-agent trajectories while enhancing numerical stability during training, enabling more effective use of scene conditioning than cross-attention alone.
        
		During post-training, we develop a stable online RL post-training framework to strengthen safety- and efficiency-oriented cooperative behaviors. We formulate a \emph{two-layer MDP} to support online optimization, explicitly coupling the denoising chain with policy--environment interaction. To counteract training instability induced by closed-loop interaction, we design dense and well-shaped rewards to characterize cooperative behaviors with respect to safety and efficiency, providing stable and fine-grained optimization signals. We further propose \emph{variance-gated group-relative policy optimization (VG-GRPO)}, which adaptively gates sampled groups and switches normalization schemes based on within-group reward variance, mitigating advantage collapse and gradient instability in standard GRPO~\cite{guo2025deepseek} and improving the robustness of online training.

		\noindent\textbf{Our main contributions are summarized as follows:}
		\begin{itemize}
        
        \item We propose \textsc{SCORP}, a tightly coupled framework that unifies scene-consistent multi-agent diffusion pre-training with stable online RL post-training for cooperative driving. The pre-training stage learns a scene-conditioned multimodal interaction prior that provides realistic, diverse grouped trajectory samples for closed-loop exploration and group-relative policy optimization. Building on this prior, the post-training stage performs online refinement to progressively promote safety- and efficiency-oriented cooperative behaviors while preserving scene consistency and interaction realism.

        \item  We propose a scene-conditioned multi-agent trajectory diffusion model that couples inter-agent self-attention with a dual-path conditioning mechanism, thereby jointly capturing inter-agent interactions and improving the scene consistency and road adherence of joint trajectories.
        Meanwhile, we develop a stable online RL post-training pipeline that explicitly integrates the denoising chain with policy--environment interaction through a two-layer MDP formulation. To stabilize closed-loop training, we co-design dense, well-shaped rewards and VG-GRPO, thereby mitigating advantage collapse and gradient instability while achieving continual improvement in closed-loop planning and cooperative behavior.

		\item Extensive experiments demonstrate that SCORP achieves superior closed-loop planning performance, yielding 10.47\%--28.26\% and 1.70\%--7.22\% improvements in safety and efficiency metrics, respectively, over strong representative baselines on WOMD~\cite{ettinger2021large}. Furthermore, compared with other post-training methods, the proposed online RL framework delivers larger and more consistent gains across key closed-loop metrics.  

		\end{itemize}

        The remainder of this paper is organized as follows. Section \ref{sec:related_work} reviews the related work. Section \ref{sec:Preliminaries} introduces the preliminaries of diffusion models and reinforcement learning. Section \ref{sec:problem} presents the problem statement. Section \ref{sec:gen_model} details the proposed multi-agent diffusion pre-training method. Section \ref{sec:rl} presents the stable online RL post-training pipeline. Section \ref{sec:results} reports the simulation results and discussion. Finally, Section \ref{sec:conclusion} concludes this paper.
	
	\section{Related Work}\label{sec:related_work}
    This section reviews the studies most relevant to our method, focusing on multi-agent behavior modeling in traffic scenarios and reinforcement post-training for driving planners.

	\subsection{Multi-agent Behavior Modeling in Traffic Scenarios}
	
Modeling the joint behavior of multiple interacting agents is essential for advancing autonomous systems, yet it remains challenging because future motion is inherently multimodal, long-horizon, and tightly coupled across participants \cite{chen2024vadv2,nayakanti2022wayformer}.
Typical distributional regression methods fit parametric continuous distributions such as Gaussian \cite{shi2024mtr++} , Laplace \cite{zhou2023qcnext} to obtain a compact multimodal representation.
Other studies incorporate goal anchors \cite{gu2021densetnt} or learnable intention queries \cite{varadarajan2022multipath++,shi2024mtr++} into the decoding process to generate multi-modal future motions.
However, these designs often increase model complexity and memory usage, limiting scalability.

Recently, generative models, notably autoregressive Transformer and diffusion models, have advanced multi-agent planning and simulation \cite{wu2024smart,seff2023motionlm,huang2025gen}.
Autoregressive Transformer models cast multi-agent motion modeling as a next-token prediction task \cite{zhou2024behaviorgpt,yuan2021agentformer}, with methods such as SMART \cite{wu2024smart} learning categorical distributions over discrete motion tokens, and MotionLM \cite{seff2023motionlm} enabling weighted mode identification via pairwise sampling and a simple rollout aggregation. However, their discrete, sequential decoding can hinder temporal coherence and remain limited to partially joint dependencies.

By contrast, diffusion models offer a compelling alternative, as they excel at modeling complex multimodal distributions while producing temporally consistent trajectories. They have been applied to decision-making, trajectory planning, and traffic simulation \cite{jiang2024scenediffuser,jiang2023motiondiffuser,zhong2023guided}. For example, CTG++\cite{zhong2023language} employs a spatiotemporal Transformer that captures the evolving dynamics of multi-agent interactions, and VBD \cite{huang2024versatile} combines a diffusion trajectory model with behavior prediction to produce versatile traffic behaviors. Despite improved behavioral diversity, existing diffusion-based methods often struggle to balance agent interaction modeling with scene-conditioned consistency, leading to scene-inconsistent multi-agent trajectories \cite{huang2024versatile}. Moreover, these methods are susceptible to closed-loop distribution shift and objective misalignment, limiting robustness and reliability in real-world deployment \cite{li2026plannerrft}. In contrast, our method is built around this gap: \textit{we retain diffusion’s multimodal expressiveness while explicitly coupling scene-conditioned pre-training with RL post-training to optimize closed-loop {safety} and {efficiency}.}
	
	\subsection{Reinforcement Post-training for Driving Planners}
	
Reinforcement learning (RL) has been shown to further enhance the capabilities of pretrained models by combining sampling-based exploration with reward-driven policy optimization \cite{zhang2025carplanner,jiang2025alphadrive,li2026plannerrft}.
Recent studies on reinforcement learning for driving planners generally focus on fine-tuning two major classes of generative models.
The first line of work, analogous to RL fine-tuning for large language models, uses autoregressive generation to model each motion token as a continuous distribution and perform policy improvement \cite{li2024think2drive,peng2024improving,tang2025plan}.
However, this approach inherently suffers from conflicts between sequence-level and token-level objectives, and temporal instability induced by sequential decoding.
In contrast, diffusion models provide an inherently temporally consistent decision process and produce diverse actions through probabilistic denoising in continuous space, making them particularly well-suited to RL’s exploration and exploitation paradigm.
For example, TrajHF\cite{li2025finetuning} proposes a human feedback-driven RL fine-tuning scheme that aligns generative trajectory models with diverse human driving preferences. ReCogDrive\cite{li2025recogdrive} fine-tunes the diffusion planner using RL with a non-reactive simulator to generate safer and more stable trajectories.

Nevertheless, existing methods are predominantly offline and do not interact with reactive environments; consequently, limited rollouts and exploration often yield marginal performance gains \cite{li2025finetuning,li2025recogdrive,huang2025gen}.
More importantly, RL training in fully closed-loop settings remains challenging due to non-stationary multi-agent interactions, compounding rollout errors, and high-variance gradient estimates arising from closed-loop environment interaction \cite{gao2025rad,li2025recogdrive}.
Taken together, existing diffusion-based multi-agent planners typically leave two key gaps unresolved: insufficient scene-conditioned denoising and predominantly offline post-training. \textit{Therefore, our method addresses both gaps by coupling a scene-conditioned denoising architecture with analytically tractable reverse-kernel online optimization, further stabilized by dense rewards and variance-gated updates, thereby improving closed-loop cooperative driving performance.}
	\section{Preliminaries}\label{sec:Preliminaries} 
	
	This section introduces the preliminaries and notation of scene-conditioned diffusion-based planning and reinforcement learning used throughout the paper.
	
	\subsection{Diffusion Model and Diffusion Policies}\label{sec:ddpm} 
	
	Denoising Diffusion Probabilistic Models (DDPMs)~\cite{nichol2021improved} define a forward noising process together with a learned reverse denoising process for sample generation. In our setting, the clean sample $\mathbf{u}_0$ denotes a future \emph{action chunk}, i.e., a control sequence to be generated for the controlled agents over the planning horizon, conditioned on the scene context $\mathbf{c}$. Given a clean action chunk $\mathbf{u}_0$, the forward diffusion process gradually perturbs it into Gaussian noise through a Markov chain
	\begin{equation}
		q(\mathbf{u}_{1:K}\mid \mathbf{u}_0)
		:=
		\prod_{k=1}^{K}q(\mathbf{u}_k\mid \mathbf{u}_{k-1}),
	\end{equation}
	where
	\begin{equation}
		q(\mathbf{u}_k\mid \mathbf{u}_{k-1})
		=
		\mathcal{N}\!\left(
		\mathbf{u}_k;\sqrt{1-\beta_k}\,\mathbf{u}_{k-1},\beta_k\mathbf{I}
		\right),k=1,\ldots,K.
	\end{equation}
	Here, $\beta_k\in(0,1)$ is a predefined noise schedule, $\alpha_k=1-\beta_k$, and $\bar{\alpha}_k=\prod_{i=1}^{k}\alpha_i$. The noisy variable at step $k$ admits the closed-form expression
	\begin{equation}
		\mathbf{u}_k
		=
		\sqrt{\bar{\alpha}_k}\,\mathbf{u}_0
		+
		\sqrt{1-\bar{\alpha}_k}\,\boldsymbol{\epsilon},
		\qquad
		\epsilon\sim\mathcal{N}(0,\mathbf{I}).
	\end{equation}
	The forward diffusion process is independent of the scene context $\mathbf{c}$ once $\mathbf{u}_0$ is given.
	
	Starting from Gaussian noise $\mathbf{u}_K\sim\mathcal{N}(0,\mathbf{I})$, the reverse process generates actions conditioned on ${\mathbf{c}}$:
	\begin{equation}
		p_\theta(\mathbf{u}_{0:K}\mid {\mathbf{c}})
		:=
		p(\mathbf{u}_K)\prod_{k=1}^{K}p_\theta(\mathbf{u}_{k-1}\mid \mathbf{u}_k,{\mathbf{c}}),
	\end{equation}
	where each reverse transition is parameterized as a Gaussian
	\begin{equation}\label{se05:denoised_sample}
		p_\theta(\mathbf{u}_{k-1}\mid \mathbf{u}_k,{\mathbf{c}})
		:=
		\mathcal{N}\!\left(
		\mathbf{u}_{k-1};
		\mu(\mathbf{u}_k,\mathcal{D}_{\theta}(\mathbf{u}_k,{\mathbf{c}},k)),
		\sigma_k^2\mathbf{I}
		\right).
	\end{equation}
	Here, $\mathbf{u}_k$ denotes the noisy action variable at denoising step $k$, and $\mathbf{u}_{k-1}$ is its one-step denoised version. The reverse mean $\mu(\cdot)$ is computed by the standard DDPM closed-form expression from the current noisy action chunk $\mathbf{u}_k$ and the denoiser output $\mathcal{D}_{\theta}(\mathbf{u}_k,\mathbf{c},k)$, while $\sigma_k^2$ is determined by the fixed diffusion schedule. After $K$ reverse steps, the final action chunk $\mathbf{u}_0$ is obtained. The corresponding conditional diffusion policy $\pi_\theta(\mathbf{u}_0\mid {\mathbf{c}})$ is defined as the marginal distribution over $\mathbf{u}_0$ induced by the conditioned reverse denoising chain.

    \subsection{Markov Decision Process and Policy Optimization}\label{sec:mdp}

A Markov decision process (MDP) is defined by a tuple
\(\mathcal{M}=(\mathcal{S},\mathcal{A},P_0,P,R)\),
where \(\mathcal{S}\) and \(\mathcal{A}\) denote the state and action spaces, \(P_0\) is the initial-state distribution, \(P\) is the transition kernel, and \(R\) is the immediate reward function. At each time step \(t\), the agent observes a state \(s_t\in\mathcal{S}\), samples an action \(a_t\in\mathcal{A}\) from the policy \(\pi_\theta(\cdot\mid s_t)\), receives reward \(R(s_t,a_t)\), and transitions to the next state according to \(s_{t+1}\sim P(\cdot\mid s_t,a_t)\).

Starting from \(s_0\sim P_0\), the policy together with the environment dynamics induces a trajectory distribution over state--action sequences, i.e.,
\(\tau=(s_0,a_0,s_1,a_1,\ldots)\).
The RL objective is to maximize the expected discounted return
\begin{equation}
	\mathcal{J}(\pi_\theta)
	=
	\mathbb{E}_{\tau\sim(\pi_\theta,P_0,P)}
	\left[
	\sum_{t=0}^{\infty}\gamma^t R(s_t,a_t)
	\right],
\end{equation}
where \(\gamma\in(0,1)\) is the discount factor.

Policy-gradient methods optimize this objective using estimators of the form
\begin{equation}
	\nabla_\theta \mathcal{J}(\pi_\theta)
	=
	\mathbb{E}_{\tau\sim(\pi_\theta,P_0,P)}
	\left[
	\sum_{t=0}^{\infty}
	\nabla_\theta \log \pi_\theta(a_t\mid s_t)\,G_t
	\right],
\end{equation}
where
\begin{equation}
	G_t:=\sum_{\tau\ge t}\gamma^{\tau-t}R(s_\tau,a_\tau)
\end{equation}
is the \textit{return-to-go} from time \(t\).

In this paper, this generic RL formulation is instantiated through a diffusion-policy optimization problem. The executable control is not sampled in one step; instead, it is generated through a stochastic reverse denoising chain. Accordingly, in Sec.~\ref{se05_sect:HoM} we reformulate policy optimization as a two-layer MDP, where the inner MDP models the denoising process and the outer MDP models policy--environment interaction.
	
	\section{Problem Statement}\label{sec:problem}

\begin{figure}[!t]
		\centering
		\includegraphics[width = 0.98\columnwidth]{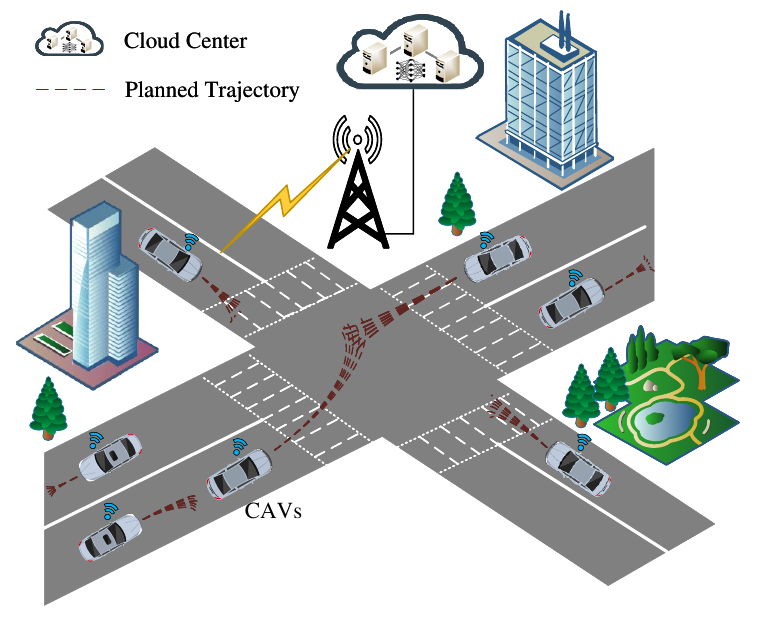}
		\caption{An illustration of a vehicle-cloud cooperative traffic scenario. A central cloud planner processes scene contexts and generates coordinated multi-agent trajectories using the proposed SCORP framework. Planned trajectories are transmitted to CAVs for execution.}
		\label{se05:fig:vehicle_colud}
	\end{figure}

    We consider the vehicle-cloud cooperative traffic scenario shown in Fig. \ref{se05:fig:vehicle_colud}, where a cloud center with powerful computational capability generates future cooperative vehicle trajectories, which are then transmitted to the vehicles for control execution. The core task is multi-agent planning in interactive traffic scenes. A scene is represented as $\mathcal{S}=(\mathbf{x},\mathbf{u},\mathbf{c})$, where $\mathbf{x}\in\mathbb{R}^{N_a\times T\times D_x}$ and $\mathbf{u}\in\mathbb{R}^{N_a\times T\times D_u}$ denote the state and control trajectories of $N_a$ agents over a finite horizon $T$, respectively. $D_x$ and $D_u$ denote the corresponding feature dimensions. The scene context is given by $\mathbf{c}=(\mathbf{c}_a,\mathbf{c}_{mp},\mathbf{c}_{tl})$, consisting of ($i$) \textit{agent history} $\mathbf{c}_a$, ($ii$) \textit{lane graph} $\mathbf{c}_{mp}$, and ($iii$) \textit{traffic-light states} $\mathbf{c}_{tl}$. 
    We formulate multi-vehicle trajectory planning in a traffic scene as a future trajectory generation task, where planning corresponds to sampling from a target distribution.
	Specifically, we aim to learn a scene-conditioned \emph{joint} planning policy $\pi_\theta(\cdot\mid\mathbf{c})$ that produces interaction-realistic multi-vehicle behaviors while improving closed-loop \textit{safety} and \textit{efficiency}.  
	
	To this end, leveraging diffusion models’ strong expressiveness for complex distributions, we first train a multi-agent trajectory diffusion model via imitation learning to capture multimodal interactive behaviors, which jointly generate future trajectories for all vehicles conditioned on the scene context $\mathbf{c}$.
	However, pretrained diffusion policies $\pi_\theta$ often suffer from distribution shift and objective misalignment. A key question is: \emph{how can we preserve realistic interactive trajectories while promoting safety- and efficiency-oriented cooperative driving?}

    Therefore, we leverage reinforcement learning post-training to improve closed-loop planning performance and strengthen cooperative behavior. Specifically, we sample rollouts $\mathbf{x}_0\sim \pi_\theta(\cdot\mid \mathbf{c})$ and optimize $\pi_\theta$ by maximizing the expected cumulative reward $\mathcal{J}(\pi_\theta)$.
    Finally, we propose \textsc{SCORP}, a scene-consistent cooperative multi-agent planner that couples condition-enhanced diffusion pre-training with stable online RL post-training.
    
	\section{Scene-consistent Multi-Agent Diffusion Pre-Training}\label{sec:gen_model}
	
\begin{figure*}[!t]
		\centering
		\includegraphics[width=1\textwidth]{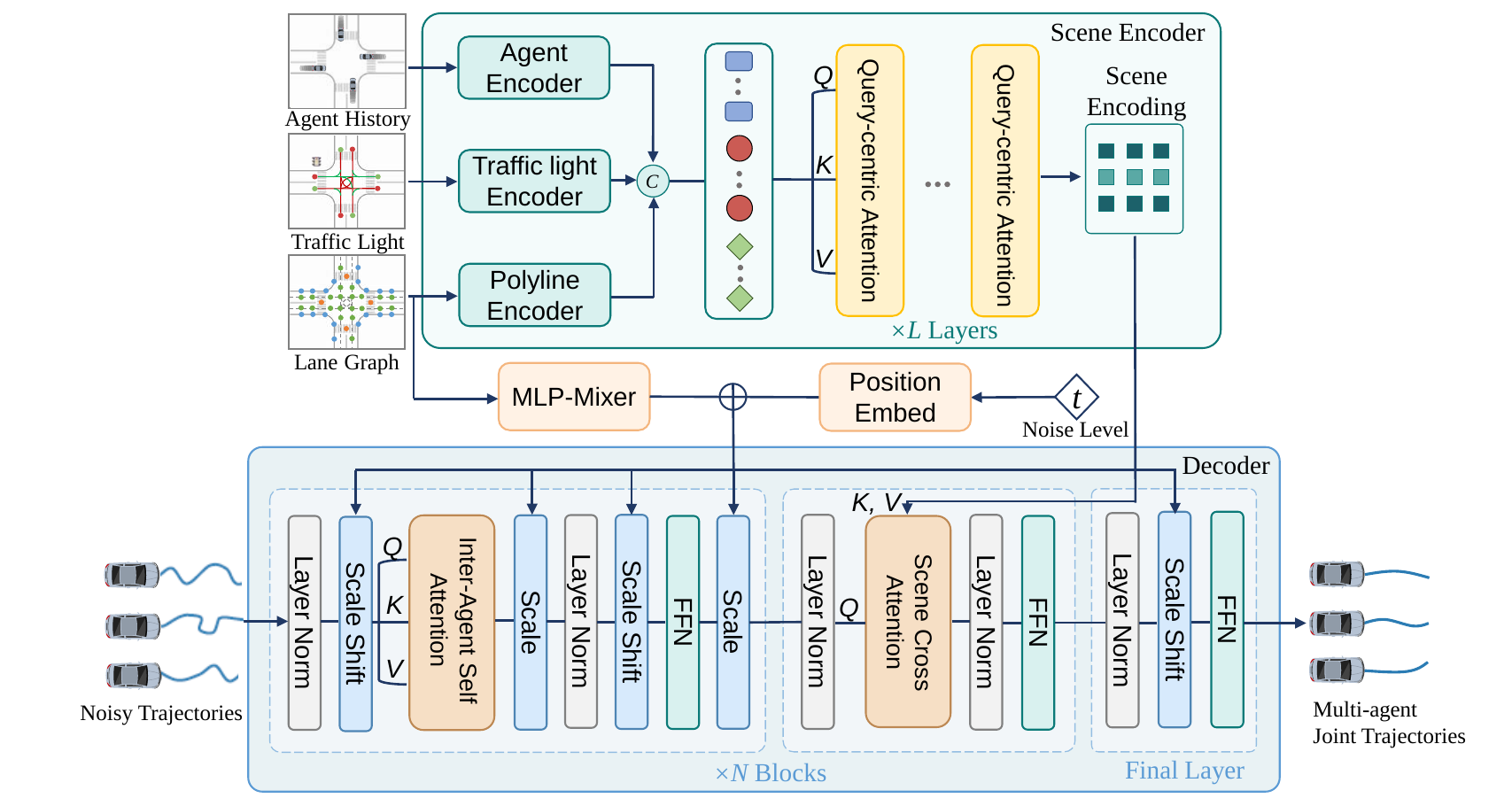}
		\caption{Architecture of the multi-agent diffusion planner. A symmetric scene encoder models scene elements and their relations in local coordinates through query-centric self-attention. A denoising decoder predicts joint future plans via inter-agent self-attention, scene-conditioned generation, and AdaLN-Zero modulation.}
		\label{fig:madp}
	\end{figure*}

This section describes the scene-conditioned multi-agent diffusion planner, the pre-training half of \textsc{SCORP}. It addresses the scene-consistency gap in prior diffusion planners, while Sec.~\ref{sec:rl} leverages the analytically tractable reverse kernel for stable online optimization. To better balance inter-agent interaction modeling with scene conditioning, we combine cross-attention with AdaLN-Zero modulation to improve the scene consistency and constraint adherence of joint trajectories. We first outline the planner architecture in Fig.~\ref{fig:madp}, and then detail (i) a symmetric scene encoder for comprehensive scene conditioning and (ii) a multi-agent denoising decoder for scene-level interaction modeling.

	\subsection{Model Structure of the Multi-agent Diffusion Planner}\label{sec:arch}

	The planner $\mathcal{P}_\theta\triangleq(\mathcal{E}_{\theta_1},\mathcal{D}_{\theta_2})$ is composed of a symmetric scene encoder $\mathcal{E}_{{\theta_1}}$ and a multi-agent denoising decoder $\mathcal{D}_{\theta_2}$, as shown in Fig.~\ref{fig:madp}: 
	
	($i$)
	\textit{Symmetric scene encoder} $\mathcal{E}_{\theta_1}: \mathbf{c} \mapsto \tilde{\mathbf{c}}$. 
	The scene context is denoted by
	\(\mathbf{c}=(\mathbf{c}_a,\mathbf{c}_{mp},\mathbf{c}_{tl})\),
	where \(\mathbf{c}_a\), \(\mathbf{c}_{mp}\), and \(\mathbf{c}_{tl}\) represent agent history, lane-graph polylines, and traffic-light states, respectively. 
	The encoder uses a query-centric attention Transformer to map the raw scene context into latent scene tokens \(\tilde{\mathbf{c}}\). Specifically, scene elements are first transformed into local coordinate frames, and query-centric attention then combines token features with relative geometry to symmetrically capture pairwise relations among heterogeneous scene elements~\cite{shi2024mtr++}. Detailed descriptions are provided in Sec.~\ref{sec:encoder}.
	
	($ii$) \textit{Multi-agent denoising decoder} $\mathcal{D}_{\theta_2}:({\mathbf{u}}_k, \tilde{\mathbf{c}}, k) \mapsto \hat{\mathbf{u}}_0$, where $\hat{\mathbf{u}}_0$ is the predicted clean action chunk $\mathbf{u}_0$. Given the encoded scene context $\tilde{\mathbf{c}}$, a noisy action chunk $\mathbf{u}_k$, and the denoising step index $k$, the decoder $\mathcal{D}_{\theta_2}$ produces a denoising predicted output $\hat{\mathbf{u}}_0$. The decoder first models inter-agent dependencies through self-attention over agent trajectory tokens. It then injects scene information through cross-attention and further modulates the hidden states using AdaLN-Zero, allowing scene conditions to influence the denoising process at multiple layers. This conditioning strategy improves the consistency between generated trajectories and the surrounding scene context. Details of the decoder are given in Sec.~\ref{sec:decoder}.
	
During pre-training, the scene encoder $\mathcal{E}_{{\theta_1}}$ and the denoising decoder $\mathcal{D}_{\theta_2}$ are optimized jointly in an end-to-end manner. Algorithm~\ref{alg:pretrain_madp} summarizes the loop: sample an expert trajectory and scene context, recover the clean control chunk through inverse dynamics, add noise at a random diffusion step, denoise it conditioned on the encoded scene, roll out the predicted controls through the vehicle dynamics, and optimize a Smooth-$\ell_1$ loss in trajectory space. The procedure clarifies that diffusion is defined in action space while supervision is imposed on the induced multi-agent trajectories.

\begin{algorithm}[!t]
    \fontsize{9.5pt}{11pt}\selectfont
    \DontPrintSemicolon 
	\caption{Pre-training of the multi-agent diffusion planner}
	\label{alg:pretrain_madp}
	\KwIn{Planner $\mathcal{P}_{\theta}=(\mathcal{E}_{\theta_1},\mathcal{D}_{\theta_2})$, dataset $\mathcal{D}_{\mathrm{pre}}$, diffusion steps $K$, dynamics $f(\cdot)$.}
	
	\For{each training iteration}{
		Sample expert trajectory $\mathbf{x}_{gt}$ and scene context $\mathbf{c}$ from $\mathcal{D}_{\mathrm{pre}}$;\;
		Get action trajectory $\mathbf{u}_0 \leftarrow f^{-1}(\mathbf{x}_{gt})$;\;
		Sample diffusion step $k \sim \mathcal{U}\{1,\ldots,K\}$ and Gaussian noise $\epsilon \sim \mathcal{N}(0,\mathbf{I})$;\;
		Form noisy action trajectory $\mathbf{u}_k \leftarrow \sqrt{\bar{\alpha}_k}\mathbf{u}_0 + \sqrt{1-\bar{\alpha}_k}\epsilon$;\;
		Predict denoised action chunk $\hat{\mathbf{u}}_0 \leftarrow \mathcal{D}_{\theta_2}(\mathbf{u}_k, \mathcal{E}_{\theta_1}(\mathbf{c}), k)$;\;
		Roll out denoised trajectory $\hat{\mathbf{x}}_0 \leftarrow f(\hat{\mathbf{u}}_0)$;\;
		Compute loss $\mathcal{L}_{\theta} \leftarrow \mathcal{SL}_1(\hat{\mathbf{x}}_0-\mathbf{x}_{gt})$;\;
		Update $\theta=(\theta_1,\theta_2)$ by backpropagation;\;
	}
	\Return{Converged parameters $\theta$}\;
\end{algorithm}

	\subsection{Symmetric Scene Context Encoding}\label{sec:encoder}
	
	\textbf{Tokenization.}
	The scene context $\mathbf{c}$ consists of three modalities: agent history $\mathbf{c}_a \in \mathbb{R}^{N_a \times T_h \times D_a}$, lane graph polylines $\mathbf{c}_{mp} \in \mathbb{R}^{N_m \times M_w \times D_p}$, and traffic-light states $\mathbf{c}_{tl} \in \mathbb{R}^{N_t \times D_t}$, where $N_a$, $N_m$, and $N_t$ denote the numbers of agents, map polylines, and traffic lights, respectively; $T_h$ is the number of observed historical steps; $M_w$ is the number of waypoints in each polyline; and $D_a$, $D_p$, and $D_t$ denote the corresponding feature dimensions. For each agent, the historical states are embedded by an MLP and fused with a learnable agent-type embedding to form an agent token.     For each map polyline, waypoint-level geometric attributes are first encoded by an MLP and then aggregated by max pooling along the waypoint dimension to obtain a polyline-level feature, which is further fused with a discrete type embedding to form a polyline token. For each traffic light, the stop-point coordinates and signal state are encoded by an MLP to produce a traffic-light token. In this way, heterogeneous scene elements are converted into a unified token representation.

	\textbf{Local coordinates and query-centric attention.}
	Before applying the Transformer encoder, positional attributes are transformed into local coordinate systems. Specifically, each agent is represented relative to its last observed state, and each polyline is represented relative to its first waypoint. This normalization reduces the burden of modeling absolute positions and emphasizes the relative geometric structure of the scene. The token set is then processed by $L$ query-centric Transformer layers following~\cite{shi2024mtr++}. For a query token $i$ at layer $\ell$, let $P_i^{(\ell)} \in \mathbb{R}^{D}$ denote its embedding, and let $\Omega(i)$ denote the set of neighboring tokens associated with token $i$. For each neighbor $j \in \Omega(i)$, we express its pose in the local coordinate frame of token $i$ and compute the relative geometric descriptor
	$R_{ij}=(\Delta x_{ij},\Delta y_{ij},\Delta\psi_{ij})$. Let $\mathrm{PE}(R_{ij})$ be a relative positional encoding and let $[\cdot,\cdot]$ denote the concatenation. The query-centric self-attention is
	\begin{equation}
		\begin{aligned}
			&P_i^{\prime(\ell)}=
			\mathrm{MHSA}\Big(
			\mathrm{Q}:\big[P_i^{(\ell)},\mathrm{PE}(R_{ii})\big],\;\\&
			\mathrm{K}:\big\{\big[P_j^{(\ell)},\mathrm{PE}(R_{ij})\big]\big\}_{j\in\Omega(i)},\;
			\mathrm{V}:\big\{P_j^{(\ell)}+\mathrm{PE}(R_{ij})\big\}_{j\in\Omega(i)}\Big),
		\end{aligned}
	\end{equation}
	where MHSA($\cdot_\text{query}$, $\cdot_\text{key}$, $\cdot_\text{value}$) denotes the multi-head self-attention operator.
	We apply the same computation to all tokens and produce the unified scene encoding, denoted by $\tilde{\mathbf{c}} \in \mathbb{R}^{(N_a+N_m+N_t) \times D}$.
    
	\subsection{Multi-agent Trajectory Denoising Decoding}\label{sec:decoder}

    We define diffusion over the joint control chunks of vehicles in the action space and learn a denoiser $\mathcal{D}_{\theta_2}$ conditioned on the scene encoding $\tilde{\mathbf{c}}$ and diffusion step $k$. The noisy controls are then rolled out through the dynamics $f(\cdot)$ to obtain corresponding noisy state trajectories, which expose vehicle motion and inter-agent interactions more explicitly and better match the trajectory-level supervision used in training. These trajectories are encoded by an MLP and fed into the denoiser, which first models multi-agent dependencies through Transformer blocks with inter-agent self-attention.
	

	\textbf{Scene-conditioned modeling via cross-attention and AdaLN-Zero.}
	We propose a dual-path scene conditioning mechanism to model the scene-consistent joint trajectory distribution\footnote{In multi-agent denoising with strong inter-agent interactions, relying on cross-attention alone can underutilize scene context, especially when road-graph features are relatively sparse, thereby increasing the risk of scene inconsistency and scene-constraint violations~\cite{huang2024versatile}.}. As shown in Fig.~\ref{fig:madp}, we couple (i) \emph{cross-attention}, which fuses the scene encoding as keys/values to update trajectory tokens, with (ii) \emph{AdaLN-Zero}, which modulates trajectory tokens conditioned on the diffusion timestep and a dense road-feature representation. The former provides a direct information-injection pathway from the scene encoder, while the latter provides a flexible and stable constraint-enforcement pathway that remains active throughout the denoising stack. 
	Specifically, since the road-graph embedding $\tilde{\mathbf{c}}_{mp}$ can be relatively sparse, we first densify it with a lightweight MLP-Mixer block. Here, $\mathrm{MLP}_{\text{tok}}$ denotes a token-mixing MLP that propagates information across tokens, and $\mathrm{MLP}_{\text{ch}}$ denotes a channel-mixing MLP that mixes feature dimensions. The update is written as
	\begin{equation}
		\begin{aligned}
			&\tilde{\mathbf{c}}_{mp}\leftarrow \tilde{\mathbf{c}}_{mp}+\mathrm{MLP}_{\text{tok}}\!\left(\tilde{\mathbf{c}}_{mp}^{\top}\right)^{\top},\\
			&\tilde{\mathbf{c}}_{mp}\leftarrow \tilde{\mathbf{c}}_{mp}+\mathrm{MLP}_{\text{ch}}\!\left(\tilde{\mathbf{c}}_{mp}\right).
		\end{aligned}
	\end{equation}    
	AdaLN then modulates each trajectory-token feature $\mathbf{x}$ by
	\begin{equation}
		\mathrm{AdaLN}(\mathbf{x})=\big(1+\gamma(k,\tilde{\mathbf{c}}_{mp})\big)\odot \mathbf{x}+\beta(k,\tilde{\mathbf{c}}_{mp}),
	\end{equation}
	where $\gamma(\cdot)$ and $\beta(\cdot)$ are the scale and shift factors regressed from the sum of the diffusion-step embedding and the dense road features.
	We further regress a gating factor $\alpha(k,\tilde{\mathbf{c}}_{mp})$ applied to the residual branch. 
	We initialize all $\alpha$ to zero so that the full decoder starts as an identity function and gradually learns to introduce conditioning, which strengthens conditional modulation and improves training stability.

	Finally, the noise embedding is propagated to the final layer, which outputs the denoised action sequence $\mathbf{u}_0$. We train the encoder $\mathcal{E}_{\theta_1}$ and the denoising decoder $\mathcal{D}_{\theta_2}$ by minimizing the loss:
	\begin{equation}\label{diffuion_cost}
		\mathcal{L}_{\theta}=\mathbb{E}_{\mathbf{u}_0,\! \; k ,\! \; \mathbf{u}_k \sim p_k(\cdot \mid \mathbf{u}_0)}\!
		\left[
		\mathcal{SL}_1\!\left(\mathbf{x}_0\!\left(\mathcal{D}_{\theta_2}(\mathbf{u}_k, \mathcal{E}_{\theta_1}\!({\mathbf{c}}), k)\right)\!-\!\mathbf{x}_{gt}\right)
		\right],
	\end{equation}
	where $\mathcal{SL}_1(\cdot)$ is the Smooth-$\ell_1$ loss between the expert trajectory $\mathbf{x}_{gt}$ and the trajectory $\mathbf{x}_0$ induced by executing the denoised controls $\mathbf{u}_0$ through the dynamics $f(\cdot)$.

	
	\section{Stable Online Reinforcement Learning Post-Training}\label{sec:rl}

    A multi-agent diffusion planner trained via imitation learning can capture general driving capabilities, but it often degrades under closed-loop execution due to distribution shift and the scarcity of safety-critical interaction data. Moreover, it fails to explicitly promote safety- and efficiency-oriented cooperative behaviors, relying solely on the BC loss.

    To address these limitations, we develop a stable \emph{online} RL post-training framework for closed-loop cooperative driving, as shown in Fig.~\ref{se05:fig:RL_training}. We first formulate a two-layer MDP to enable online optimization. We then co-design dense, well-shaped rewards and VG-GRPO to stabilize online training during closed-loop interaction. Finally, we present the overall RL post-training pipeline.

	\subsection{Two-layer MDP Formulation}\label{se05_sect:HoM}
	
\begin{figure}[!t]
		\centering
		\includegraphics[width = 0.98\columnwidth]{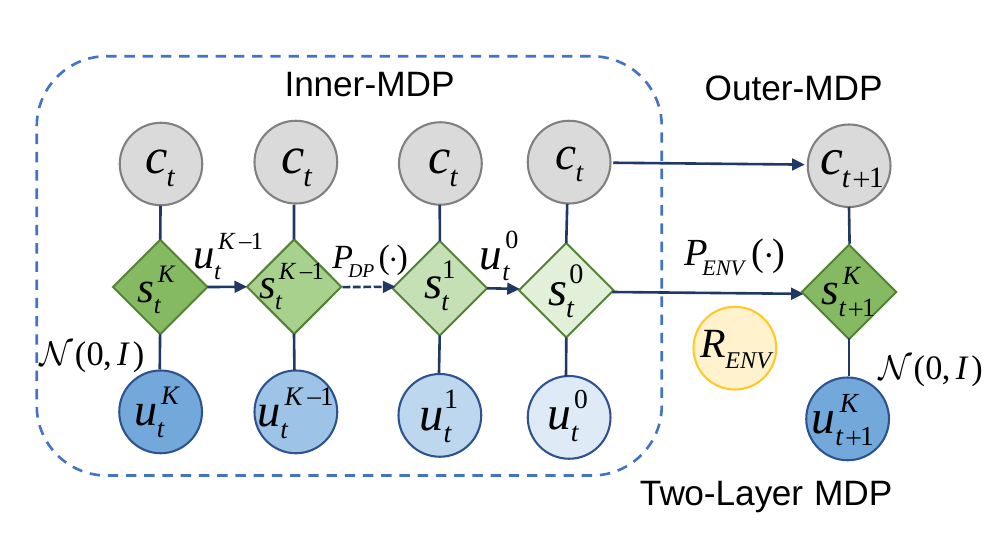}
		\caption{Two-layer MDP formulation for reinforcement learning. We construct a two-layer MDP consisting of an inner denoising MDP and an outer environment-interaction MDP, and use Gaussian likelihoods at each denoising step so that the denoising process can be optimized via policy gradients.}
		\label{se05:fig:mdp_formulation}
	\end{figure}

	We now tailor the MDP to diffusion-policy optimization. Unlike standard policies that sample $a_t$ in one shot, a diffusion policy produces the executed action $\mathbf{u}_t^0$ through a \emph{$K$-step stochastic denoising chain}. Treating the entire denoising chain as a black-box sampler in a single-level MDP leads to poor credit assignment across denoising steps and high-variance gradients, which is exacerbated in \emph{multi-vehicle online} rollouts due to severe non-stationarity from coupled agent interactions. We therefore model denoising as an inner MDP $\mathcal{M}_{\mathrm{DP}}$ with \emph{analytically tractable Gaussian likelihoods} at each step, coupled with an outer environment-interaction MDP $\mathcal{M}_{\mathrm{ENV}}$ that provides closed-loop rewards and constraints. Following~\cite{black2023training}, their composition yields the two-layer MDP in Fig.~\ref{se05:fig:mdp_formulation}. We use subscript $t$ for the outer environment step and superscript $k$ for the inner denoising step.
	\begin{itemize}
		\item \textbf{State.} $\bar{\mathbf{s}}_t^k=(\mathbf{c}_t,\mathbf{u}_t^k)$, where $\mathbf{c}_t$ is the scene observation at outer time step $t$ and $\mathbf{u}_t^k$ is the intermediate denoising output at denoising step $k$. 
		\item \textbf{Action.} $\bar{\mathbf{a}}_t^k=\mathbf{u}_t^{k-1}$, sampled from the step-wise reverse transition $p_{\theta}(\mathbf{u}_t^{k-1}\mid \mathbf{u}_t^{k}, \mathbf{c}_t)$; the final denoised action is executed in the outer MDP.		
		\item \textbf{Initial distribution.}  $\bar{\mathbf{s}}_0^K=(\mathbf{c}_0,\mathbf{u}_0^K)\sim\bar{\mathbf{P}}_0$, where $\mathbf{c}_0$ is drawn from the outer MDP’s initial distribution and $\mathbf{u}_{0}^K \sim \mathcal{N}(0,\mathbf{I})$.
		\item \textbf{Transition.} The kernel $\bar{\mathbf{P}}(\bar{\mathbf{s}}_{t'}^{k'}\mid \bar{\mathbf{s}}_{t}^k,\bar{\mathbf{a}}_{t}^k)$ is defined as
		\begin{equation}
			\bar{\mathbf{s}}_{t'}^{k'}\!=\!
			\left\{
			\begin{array}{ll}
				\!\!\!(\mathbf{c}_t,{\mathbf{u}}_t^{k-1}) & (t',k')=(t,k-1),\ k>0,\\[2pt]
				\!\!\!(\mathbf{c}_{t+1},\mathbf{u}_{t+1}^{K}) & (t',k')=(t\!+\!1,K),\ k=0,
			\end{array}
			\right.
		\end{equation}
		For $k > 0$, transitions occur within the inner denoising chain: $\mathbf{c}_{t}$ is fixed and the action is updated to $\mathbf{u}_{t}^{k-1}$.
		For $k=0$, we execute $\mathbf{u}_t^0$ in the environment to obtain $\mathbf{c}_{t+1}$ and re-initialize $\mathbf{u}_{t+1}^K\sim\mathcal{N}(0,\mathbf{I})$.
		\item \textbf{Reward.} $\bar{\mathbf{R}}(\bar{\mathbf{s}}_t^k,\bar{\mathbf{a}}_t^k)$ is assigned only when denoising reaches $k=0$:
		\begin{equation}
			\bar{\mathbf{R}}(\bar{\mathbf{s}}_t^k,\bar{\mathbf{a}}_t^k)=
			\left\{
			\begin{array}{ll}
				0 & k>0,\\
				R_{\mathrm{ENV}}(\mathbf{c}_t,\mathbf{u}_t^0) & k=0,
			\end{array}
			\right.
		\end{equation}
		where $R_{\mathrm{ENV}}$ is specified in Sec.~\ref{sec:reward_model}.
		\item \textbf{Step-wise likelihood.} In the inner MDP, the policy is identified with the reverse diffusion kernel. $\bar{\pi}_{\theta}(\bar{\mathbf{a}}_t^k\mid \bar{\mathbf{s}}_t^k)$ can be evaluated by computing the diffusion-policy likelihood of each denoising step along the sampled denoising chain.
		\begin{equation}\label{likelihood_pro}
			\begin{aligned}
				&\bar{\pi}_{\theta}(\bar{\mathbf{a}}_t^k\mid \bar{\mathbf{s}}_t^k)
				=p_\theta(\mathbf{u}_t^{k-1}\mid \mathbf{u}_t^{k},\mathbf{c}_t)\\
				&=\mathcal{N}\!\left(\mathbf{u}_t^{k-1};\mu\left(\mathbf{u}_t^{k}, \mathcal{D}_{\theta_2}\left(\mathbf{u}_t^{k}, \mathcal{E}_{\theta_1}\!(\mathbf{c}_t), k\right)\right), \sigma_{k}^2 \mathrm{I}\right) 
			\end{aligned}
		\end{equation}
		where $p_\theta(\mathbf{u}_t^{k-1}\mid \mathbf{u}_t^{k},\mathbf{c}_t)$ is a Gaussian transition with mean $\mu(\cdot)$ calculated from $\mathbf{u}_t^{k}$ and denoiser output $\mathcal{D}_{\theta_2}\left(\mathbf{u}_t^{k}, \mathcal{E}_{\theta_1}\!(\mathbf{c}_t), k\right)$, and variance $\sigma_k^2$ specified by the fixed noise schedule.
		Hence, the step-wise log-likelihood is analytically tractable.
		
		\item \textbf{Objective.} The objective of the two-layer MDP is
		\begin{equation}
			\bar{\mathcal{J}}\left(\bar{\pi}_\theta\right)=\mathbb{E}_{\bar{\pi}_\theta, \bar{\mathbf{P}}, \bar{\mathbf{P}}_0}\left[\sum_{t \ge 0}\sum_{\tau \ge t}\gamma^{\tau-t}\,\bar{\mathbf{R}}\left(\bar{\mathbf{s}}_\tau^k, \bar{\mathbf{a}}_\tau^k\right)\right].
		\end{equation}
	\end{itemize}
	
	\subsection{{Dense Planning Reward Design }\label{sec:reward_model}}
	
	Rewards determine both the optimization direction and the driving preference. We use simple but fine-grained rule-based rewards that generalize across diverse large-scale scenarios and follow a safety-first principle, with explicit incentives for efficiency. The overall reward is defined as
	\begin{equation}\begin{aligned}
			R_{\mathrm{ENV}}\!= \! \sum_{h}w_c\,R_{\mathrm{coll}}(h) + w_o\,R_{\mathrm{offroad}}(h) + w_e\,R_{\mathrm{eff}}(h).
		\end{aligned}
	\end{equation}
	where $h$ is the step index within the reward horizon. 
	
	\textbf{Collision.} We define the collision reward through SAT-based oriented-box overlap checking, i.e., $\text{coll\_overlaps}(\cdot)$:
	\begin{equation}
		R_{\mathrm{coll}}(h)=\mathbbm{1}[\text{coll\_overlaps}(h)] .
	\end{equation}
	
	\textbf{Off-road.} We define the off-road reward through signed distances of the four vehicle-box corners to lane boundaries, i.e., $\text{off\_road}(\cdot)$:
	\begin{equation}
		R_{\mathrm{offroad}}(h)=\mathbbm{1}[\text{off\_road}(h)] .
	\end{equation}
	
	\textbf{Efficiency.} We define the efficiency reward by measuring normalized progress along a road centerline:
	\begin{equation}
		R_{\mathrm{eff}}(h)=\max\!\left(\frac{s_{h+1}-s_{h}}{s_{\max}},\,0\right),
	\end{equation}
	where $s_h$ is the arc-length projection of the vehicle position onto the centerline and $s_{\max}$ normalizes the progress to $[0,1]$.
	In multi-vehicle rollouts, we evaluate rewards at each step over the horizon and average them across the controlled vehicles.
	This formulation provides dense and well-shaped reward evaluations that enable finer-grained discrimination among sampled trajectories, thus yielding stable learning signals.
	To avoid \emph{trajectory} dynamic-quality collapse when emphasizing safety, we use three built-in safeguards: dynamics-aware denoising with rollout through $f(\cdot)$ to ensure trajectory feasibility, KL anchoring to the pre-training prior, and short-horizon execute-then-replan in closed loop.
	
		\begin{figure}[!t] 
		\centering
		\includegraphics[width = 0.98\columnwidth] {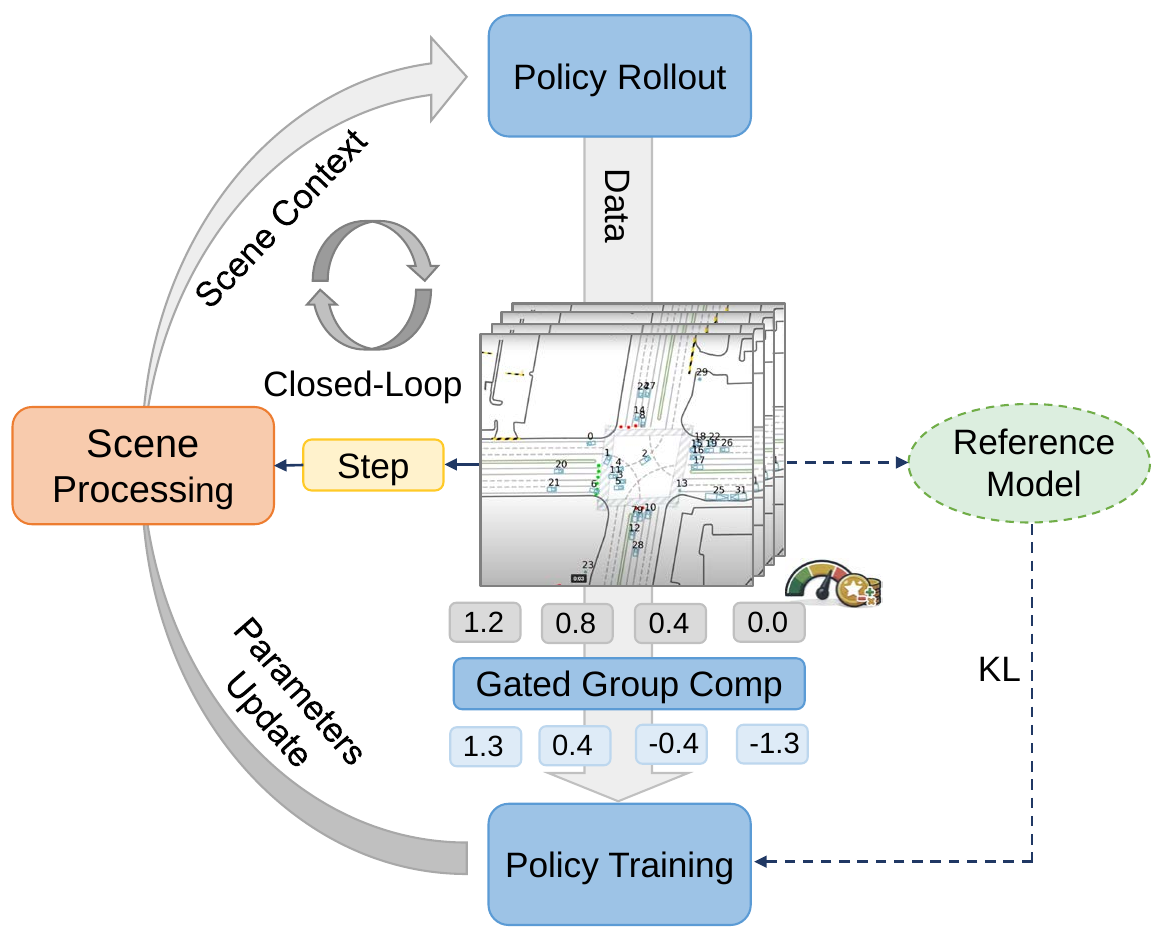} 
		\caption{Online RL post-training framework. The pipeline consists of three components: (1) closed-loop policy rollout, where the policy samples multimodal multi-agent trajectories conditioned on the updated scene context; (2) rule-based reward evaluation, where each trajectory group is scored based on safety and efficiency metrics to compute variance-gated group-relative advantages; and (3) policy training, where mini-batches are randomly sampled from the rollout buffer to optimize the actor network with policy loss and KL regularization.}
		\label{se05:fig:RL_training}
	\end{figure}
	\subsection{Variance-gated Group Relative Policy Optimization}\label{se05_sect:grpo}
	
	With the two-layer MDP (Sec.~\ref{se05_sect:HoM}) and reward (Sec.~\ref{sec:reward_model}) in place, we optimize the pretrained diffusion policy by maximizing the objective $\bar{\mathcal{J}}(\bar{\pi}_\theta)$. 
	We propose \textsc{VG-GRPO} for online refinement of the multi-agent diffusion policy. Our design is built on the critic-free GRPO paradigm~\cite{guo2025deepseek} and extends it with variance-gated advantages and denoising-aware optimization to stabilize training under multi-agent non-stationarity. This is a functional coupling rather than a loose combination: diffusion provides grouped multimodal candidates each step, and VG-GRPO directly consumes their relative ranking signal for online updates.
	
	At each outer environment step $t$, we sample a group of $G$ candidate rollouts from the old policy $\bar{\pi}_{\theta_{\mathrm{old}}}$.
	Let $i\in\{1,\ldots,G\}$ index samples in the group. Each sample yields a scalar reward $r_{t,i}:= \bar{\mathbf{R}}_{t,i}$, forming the within-group reward set $\{r_{t,1},\ldots,r_{t,G}\}$.
	
	\textbf{VG-GRPO objective.} We maximize the following objective
	\begin{equation}\label{eq:vggrpo_obj}
		\begin{gathered}
			\bar{\mathcal{J}}(\theta)=\mathbb{E}_{\big(\{\bar{\mathbf{a}}_{t,i}^k\}_{i=1}^G, \bar{\mathbf{s}}_t\big) \sim \bar{\pi}_{\theta_{o l d}}} 
			\frac{1}{G} \sum_{i=1}^G -\beta \mathbb{D}_{K L}\left(\bar{\pi}_\theta| | \bar{\pi}_{r e f}\right)\\
			+\left(\min \left( \rho_{t,i}^{k} , \operatorname{clip}\big(\rho_{t,i}^{k}, 1-\varepsilon^-, 1+\varepsilon^+\big) \right) \gamma_{\text{denoise}}^k A_{t,i} \right), 
		\end{gathered}
	\end{equation}
	where $\rho_{t,i}^{k}$ is the importance ratio, computed from step-wise likelihoods under the two-layer MDP in Sec.~\ref{se05_sect:HoM}.
	\begin{equation}
		\rho_{t,i}^{k}
		:= \frac{\bar{\pi}_\theta\left(\bar{\mathbf{a}}_{t,i}^k \mid \bar{\mathbf{s}}_t^k\right)} {\bar{\pi}_{\theta_{old}}\left(\bar{\mathbf{a}}_{t,i}^k \mid \bar{\mathbf{s}}_t^k\right)},
	\end{equation} 
	and $[\varepsilon^-,\varepsilon^+]$ is the DAPO-style clipping ratio that stabilizes training and encourages exploration, and $\gamma_{\mathrm{denoise}}^k\in(0,1]$ is a denoising-step discount factor that downweights the policy-gradient contributions of noisier steps, thereby improving training stability~\cite{ren2024diffusion}.
	To prevent mode collapse during exploration and preserve general driving capabilities, we regularize post-training with a KL term to the reference policy (set to the pretrained policy here), using the non-negative unbiased estimator~\cite{schulman2020approximating} and weighting it by $\beta$ to control the regularization strength:
	\begin{equation}
		\mathbb{D}_{K L}\left(\bar{\pi}_\theta| | \bar{\pi}_{r e f}\right)=\frac{\bar{\pi}_{ref}\left(\bar{\mathbf{a}}_{t,i}^k \mid \bar{\mathbf{s}}_t^k\right)} {\bar{\pi}_\theta\left(\bar{\mathbf{a}}_{t,i}^k \mid \bar{\mathbf{s}}_t^k\right)}-\log \frac{\bar{\pi}_{ref}\left(\bar{\mathbf{a}}_{t,i}^k \mid \bar{\mathbf{s}}_t^k\right)} {\bar{\pi}_\theta\left(\bar{\mathbf{a}}_{t,i}^k \mid \bar{\mathbf{s}}_t^k\right)}-1,
	\end{equation}

	\textbf{Variance-gated advantages.}
	Standard within-group normalization can induce training instability. In simple cases (e.g., short rollouts or near-stationary vehicles), sampled rewards become nearly identical; after GRPO’s mean–std normalization, the resulting advantages collapse toward zero, yielding vanishing policy gradients. This amplifies the sensitivity of minibatch gradients to noise and directly destabilizes training.
	To address this issue, we introduce a variance-gated mechanism, inspired by~\cite{yu2025dapo} and adapted to fine-grained selection over group-based trajectory rollouts. Let $\sigma_t\triangleq\operatorname{std}\!\left(\{r_{t,1},\ldots,r_{t,G}\}\right)$,
	the group relative advantage for sample $i$ is defined as
	\begin{equation}\label{eq:vg_adv}
		\resizebox{0.97\columnwidth}{!}{$
			A_{t,i}=
			\left\{
			\begin{array}{ll}
				\text{drop this group}, & \sigma_t\leq \mathrm{std}_{1},\\[3pt]
				r_{t,i}-\operatorname{mean}\!\left(\{r_{t,1},\ldots,r_{t,G}\}\right), & \mathrm{std}_{1}<\sigma_t\leq \mathrm{std}_{2},\\[6pt]
				\dfrac{r_{t,i}-\operatorname{mean}\!\left(\{r_{t,1},\ldots,r_{t,G}\}\right)}{\operatorname{std}\!\left(\{r_{t,1},\ldots,r_{t,G}\}\right)}, & \sigma_t>\mathrm{std}_{2},
			\end{array}
			\right.
			$}
	\end{equation}
	where $\mathrm{std}_{1}$ and $\mathrm{std}_{2}$ denote the tunable gating parameters. During policy sampling, we measure group diversity via the within-group standard deviation of rewards.
	($i$) If the within-group standard deviation is near zero, rollouts are essentially identical, so we discard the group to avoid vanishing gradients.
	($ii$) If it is small but nonzero, the differences may be noise-driven. Normalization would amplify this noise and corrupt the update direction, so we keep the raw  differences to retain sample utility while reducing instability.
	($iii$) If it is large, rollouts exhibit meaningful quality gaps and provide informative gradients; we then apply standard group-relative normalization to compute advantages.
	This variance gate retains informative samples and reduces gradient variance, thereby stabilizing typical GRPO for online RL training of multi-agent diffusion policies.
	
	\subsection{Online RL Post-training Framework}\label{se05_sect:RL_framework}
	
	We implement the online RL framework in Fig.~\ref{se05:fig:RL_training} with three stages: closed-loop policy rollout, reward evaluation, and policy training. In rollout, the policy repeatedly samples \emph{groups} of multi-agent action chunks from the latest scene context, which is refreshed after each executed action. In reward evaluation, rule-based metrics score collision, off-road behavior, and progress, producing a scalar reward for each sampled chunk. In policy update, grouped samples are used to optimize the diffusion-policy denoiser via \textbf{VG-GRPO} (Sec.~\ref{se05_sect:grpo}); the variance gate is critical for stabilizing multi-vehicle online refinement under non-stationary interactions.
	
	At each outer environment step $t$, the planner samples $G$ candidate chunks by running the inner denoising chain under the previous behavior policy $\bar{\pi}_{\theta_{\mathrm{old}}}$ conditioned on the current context $\bar{\mathbf{s}}_t^K=(\mathbf{c}_t,\mathbf{u}_t^K)$. Following~\cite{ren2024diffusion}, we use a rolling horizon: plan over $T_p$ steps but execute only the first $T_a$ steps. For group-relative optimization, we compute a scalar reward $\bar{\mathbf{R}}_{t,i}^k$ for each sample $i\in\{1,\ldots,G\}$. The environment executes the best-performing sample for interaction, while \emph{all} group samples are retained to compute variance-gated group-relative advantages and update the policy.
	
	We store grouped tuples in the rollout buffer $\mathcal{D}_{\mathrm{itr}}$:
	\begin{equation}
		\Big(\{\bar{\mathbf{s}}_{t,i}^{k}\}_{i=1}^{G},\ \{\bar{\mathbf{a}}_{t,i}^{k}\}_{i=1}^{G},\ \{\bar{\mathbf{R}}_{t,i}^k\}_{i=1}^{G},\ k\Big),
	\end{equation}
	where $\bar{\mathbf{s}}_{t,i}^{k}=(\mathbf{c}_t,\mathbf{u}_{t,i}^{k})$ and $\bar{\mathbf{a}}_{t,i}^{k}=\mathbf{u}_{t,i}^{k-1}$ follow the two-layer MDP in Sec.~\ref{se05_sect:HoM}. During training, we sample mini-batches from $\mathcal{D}_{\mathrm{itr}}$ by randomly sampling $B$ denoising steps $k$ and group indices $i$, compute the variance-gated group-relative advantages (Eq.~\eqref{eq:vg_adv}) and likelihoods (Eq.~\eqref{likelihood_pro}), and optimize the planner parameters by maximizing the VG-GRPO objective (Eq.~\eqref{eq:vggrpo_obj}).
	
	The complete post-training procedure and implementation details are summarized in Algorithm~\ref{alg:rl_posttrain_short}.
   
\begin{algorithm}[!t]
\small
\caption{Closed-loop online RL post-training}
\label{alg:rl_posttrain_short}
\LinesNumbered
\KwIn{Current policy $\bar{\pi}_\theta$, dataset $\mathcal{D}_{\mathrm{inter}}$, group size $G$, denoising steps $K$, clip $\varepsilon$, KL weight $\beta$, thresholds $\mathrm{std}_1,\mathrm{std}_2$.}

\For{each iteration}{
    $\theta_{\mathrm{old}}\leftarrow \theta$; \quad $\mathcal{D}_{\mathrm{itr}}\leftarrow \emptyset$\;
        initialize outer state $\mathbf{c}_1$ from $\mathcal{D}_{\mathrm{inter}}$\;
        \For{each outer step $t$}{
            initialize $\mathbf{u}_{t,i}^{K}\sim\mathcal{N}(0,\mathbf{I}),\ i=1,\ldots,G$\;
            set $\bar{\mathbf{s}}_{t,i}^{K}=(\mathbf{c}_t,\mathbf{u}_{t,i}^{K}),\ i=1,\ldots,G$\;
            \For{$k=K,K-1,\ldots,0$}{
                \eIf{$k>0$}{
                    sample denoising actions
                    $\bar{\mathbf{a}}_{t,i}^{k}=\mathbf{u}_{t,i}^{k-1}
                    \sim p_{\theta_{\mathrm{old}}}(\cdot\mid \mathbf{u}_{t,i}^{k},\mathbf{c}_t),\ i=1,\ldots,G$\;
                    set next inner states
                    $\bar{\mathbf{s}}_{t,i}^{k-1}=(\mathbf{c}_t,\mathbf{u}_{t,i}^{k-1})$ and $\bar{\mathbf{R}}_{t,i}^{k}=0,\ i=1,\ldots,G$\;
                }{
                    compute terminal rewards
                    $\bar{\mathbf{R}}_{t,i}^{0}\leftarrow R_{\mathrm{ENV}}(\mathbf{c}_t,\mathbf{u}_{t,i}^{0}),\ i=1,\ldots,G$\;
                    select $i^\star=\arg\max_i \bar{\mathbf{R}}_{t,i}^{0}$ and execute $\mathbf{u}_{t,i^\star}^{0}$ in the simulator to obtain $\mathbf{c}_{t+1}$\;
                }
                store
                $\Big(\{\bar{\mathbf{s}}_{t,i}^{k}\}_{i=1}^{G},
                \{\bar{\mathbf{a}}_{t,i}^{k}\}_{i=1}^{G},
                \{\bar{\mathbf{R}}_{t,i}^{k}\}_{i=1}^{G},
                k\Big)$
                in the buffer $\mathcal{D}_{\mathrm{itr}}$\;
            }
        }
    \For{each update epoch}{
        \For{each mini-batch}{
            randomly sample $B$ denoising steps $k$ and group indices $i$\;
            sample the corresponding grouped tuples $(\bar{\mathbf{s}}_{t,i}^{k}, \bar{\mathbf{a}}_{t,i}^{k}, \bar{\mathbf{R}}_{t,i}^{k}, k)$ from $\mathcal{D}_{\mathrm{itr}}$\;
            compute variance-gated advantages $A_{t,i}$ and likelihoods via Eq.~\eqref{eq:vg_adv} and \eqref{likelihood_pro}\;
            update $\bar{\pi}_\theta$ by maximizing Eq.~\eqref{eq:vggrpo_obj}\;
        }
    }
}
\Return{Converged policy $\bar{\pi}_\theta$}\;
\end{algorithm}


	
	\section{Experiments and Discussion}
	\label{sec:results}

    This section reports the closed-loop evaluation of the proposed method. We first describe the datasets, metrics, and implementation details, and then present benchmark comparisons and ablation studies. The evaluation addresses three questions: whether \textsc{SCORP} improves closed-loop safety and efficiency; whether the proposed online post-training framework yields robust gains; and how key modules and design choices, such as AdaLN-Zero, variance gating, and post-training data distribution, contribute to the overall performance.

	\subsection{Experimental Setup}
	\textbf{Dataset.}
		For pre-training, we use the Waymo Open Motion Dataset (WOMD)~\cite{ettinger2021large}, containing 486{,}995 training scenarios and 44{,}097 validation scenarios; each scenario covers 9\,s of real traffic with all participant trajectories and map topology. We perform closed-loop evaluation and ablation studies across 41{,}590 Testing Interactive scenarios. For post-training, we uniformly sample 20{,}756 scenarios from the WOMD Validation Interactive split. We then evaluate all sampled scenarios with the pretrained diffusion planner and stratify them into three subsets according to their performance scores: (i) a low-score set with 2{,}504 failure scenarios involving either inter-vehicle collisions or off-road events; (ii) a high-score set with 5{,}857 scenarios whose planned trajectories achieve high overall scores; and (iii) a full set containing all available scenarios, which mixes the low-score, high-score, and regular scenarios.
	
	\textbf{Implementation Details.}
	Scene inputs include up to $N_a\!=\!32$ agents, $N_m\!=\!256$ polylines with $M_w\!=\!30$ waypoints each, and $N_t\!=\!16$ traffic lights. The planner generates a future control sequence of length $T_p\!=\!80$ with a 0.1 s time step. We discard the entire history and condition on the current state to mitigate potential causal confusion and improve closed-loop performance. 
	In closed-loop testing, we execute $T_a\!=\!10$ steps. 
	
	The scene encoder uses $L\!=\!6$ query-centric Transformer layers with hidden size $D\!=\!256$. The denoising decoder alternates two block types for three rounds (6 Transformer layers total). MLP-Mixer token/channel dimensions are 64/128. Pre-training uses the log noise schedule in~\cite{huang2024versatile} with $\bar{\alpha}_{\min}\!=\!10^{-9}$, scaling 0.0031, and $K\!=\!20$. We pretrain the model with AdamW using a weight decay of 0.01. The learning rate is set to $2\!\times\!10^{-4}$, warmed up for 3000 steps, and decayed by a factor of 0.02 every 3000 steps.  We use gradient clipping at 1.0 and BF16 precision, and train for 30 epochs on 4 RTX 4090 GPUs with a global batch size of 32.

	Post-training uses group size $G\!=\!10$. Each rollout batch is optimized for 1 epoch with mini-batch size 16 for 10M fine-tuning steps. Following~\cite{li2026plannerrft}, we use a reward horizon of 4\,s, reward weights (collision/off-road/progress) $8/1/4$, KL weight $\beta\!=\!0.1$, and learning rate $10^{-5}$. For effective exploration, we use DAPO-style clipping $[-0.15,0.2]$~\cite{yu2025dapo}, and clamp a minimum sampling-time standard deviation of $\sigma_\text{min}^\text{sam}=0.2$. 
	For training stability, we clamp the per-step log-likelihood standard deviation to $\sigma_\text{min}^\text{prob}=0.1$ and use $\gamma_{\mathrm{denoise}}\!=\!0.9$. Moreover, we set the gating thresholds to $\mathrm{std}_1/\mathrm{std}_2=0.03/0.06$, corresponding to approximately 10\%/20\% of the empirical standard deviation of group rewards, as informed by signal-to-noise considerations~\cite{roberts2008signal} and advantage-collapse analysis~\cite{zhong2026rc}. Post-training runs in BF16 on one NVIDIA 5090 GPU.

	\textbf{Metrics.}
	We report collision rate (CR), off-road rate (OR), average speed (AS), average displacement error (ADE), and kinematic infeasibility (Kin). We treat CR, OR, and AS as \emph{primary} closed-loop objectives reflecting safety and traffic efficiency, and use ADE and Kin as \emph{secondary} diagnostics for imitation fidelity and physical feasibility. CR checks SAT-based oriented-box overlap; OR checks drivable-area boundary crossing; AS is mean per-step displacement; ADE is the $\ell_2$ position error to ground truth; and Kin counts violations of acceleration and curvature bounds with limits 6\,m/s$^2$ and 0.3\,m$^{-1}$. For fairness, all methods are evaluated with the same simulator configuration, including scenario split, agent count, horizon, and replanning frequency; values with $\pm$ report mean and standard deviation over repeated closed-loop evaluations in the main comparison tables, while ablation tables report point estimates under the same protocol for compactness.

	\subsection{Performance Evaluation}
	
	This subsection evaluates interactive trajectory generation using both quantitative benchmark results and representative qualitative cases.
	
	\begin{table}[!t]
		\caption{Closed-loop benchmark results on the WOMD testing interactive split (primary: CR, OR, AS; secondary: ADE, Kin)}\label{tab:benchmark_womd}
		\vspace{-0.5em}
		\centering\footnotesize
		\setlength{\tabcolsep}{3pt}
		\resizebox{0.99\columnwidth}{!}{%
			\begin{tabular}{lccccc}
				\toprule
				\multirow{2}{*}{\vspace{-1.4ex}Method} & \multicolumn{3}{c}{Primary closed-loop objectives} & \multicolumn{2}{c}{Secondary diagnostics} \\ \cmidrule(lr){2-4}\cmidrule(lr){5-6} & CR (\%)$\downarrow$ & OR (\%)$\downarrow$ & AS (m/s)$\uparrow$ & ADE (m)$\downarrow$ & Kin (\%)$\downarrow$ \\
				\midrule
				TrafficBotsV1.5~\cite{zhang2024trafficbots}  & 2.74$\pm$0.21  & 1.79$\pm$0.14   & 8.03$\pm$0.48  & 1.68$\pm$0.09  & 0.26$\pm$0.02 \\
				SMART-large~\cite{wu2024smart}          & 2.22$\pm$0.09  & 1.58$\pm$0.10   & 8.34$\pm$0.30  & 1.30$\pm$0.01  & \textbf{0.21$\pm$0.01} \\
				VBD~\cite{huang2024versatile}              & 2.46$\pm$0.14 & 1.92$\pm$0.18  & 8.08$\pm$0.52 & 1.41$\pm$0.02  & 0.24$\pm$0.01 \\
				SMART-tiny-CLSFT~\cite{zhang2025closed}     & 2.10$\pm$0.10  & 1.53$\pm$0.12   & 8.47$\pm$0.44  & \textbf{1.23$\pm$0.03}  & 0.25$\pm$0.02 \\
				\textsc{SCORP}           & \textbf{1.89$\pm$0.12}   & \textbf{1.36$\pm$0.08} & \textbf{8.61$\pm$0.46} & 1.36$\pm$0.04   & 0.32$\pm$0.03 \\
				\bottomrule
		\end{tabular}}
		\vspace{-0.5em}
	\end{table}
	
    \textbf{Performance Evaluation.} We compare four strong open-source baselines: VBD~\cite{huang2024versatile}, SMART-large~\cite{wu2024smart}, SMART-tiny-CLSFT~\cite{zhang2025closed}, and TrafficBotsV1.5~\cite{zhang2024trafficbots}. They cover autoregressive, behavior-cloning, and diffusion paradigms. Each scenario runs for 8 s in closed loop with replanning at 1 Hz.
	Table~\ref{tab:benchmark_womd} shows that \textsc{SCORP} leads on all three primary closed-loop objectives and delivers superior closed-loop planning performance, yielding 10.47\%--28.26\% and 1.70\%--7.22\% improvements in safety-related metrics (CR and OR) and efficiency metrics, respectively, over strong representative baselines on WOMD.
    Compared with SMART-tiny-CLSFT, a state-of-the-art open-source baseline that uses targeted closed-loop supervised fine-tuning, our method reduces CR from 2.10 to 1.89 (-10.0\%) and OR from 1.53 to 1.36 (-11.1\%), while increasing AS from 8.47 to 8.61 (+1.66\%).
	Relative to the pretrained policy in Table~\ref{tab:post_training_methods}, online post-training yields 7.4\% lower CR, 19.0\% lower OR, and 3.0\% higher AS. These results demonstrate the effectiveness of the proposed scene-conditioned pre-training and closed-loop RL post-training paradigm.

	
	\textbf{Post-training Methods.} We further compare four post-training strategies, SFT, DPO, offline RL, and online RL, as shown in Table~\ref{tab:post_training_methods}. All methods improve upon the pre-training-only policy to some extent, but with different objective alignment. SFT mainly improves imitation-oriented diagnostics (ADE). DPO and offline RL improve selected indicators but can degrade the balance between safety and efficiency. \textsc{SCORP} with online RL provides the most consistent co-improvement on the primary closed-loop objectives CR, OR, and AS, reflecting the benefit of coupling diffusion group sampling with VG-GRPO.
    
	Online RL shows a mild ADE increase, indicating a distributional trade-off: closed-loop exploration can reduce one-step imitation accuracy while improving interaction robustness. Kin remains low across methods, all below 0.4\%; compared with the pretrained policy, it increases by at most 0.07 percentage points from 0.25 to 0.32, indicating no dynamic-quality collapse under safety-oriented optimization.
    In summary, through closed-loop policy sampling, online RL continually exposes the model to more out-of-distribution scenarios, thereby improving generalization and robustness under complex interactions and offering greater potential for further gains in closed-loop performance.

	\begin{table}[!t]
		\caption{Closed-loop comparison of post-training methods (primary: CR, OR, AS; secondary: ADE, Kin)}\label{tab:post_training_methods}
		\vspace{-0.5em}
		\centering\footnotesize
		\setlength{\tabcolsep}{3pt}
		\resizebox{0.99\columnwidth}{!}{%
			\begin{tabular}{lccccc}
				\toprule
				\multirow{2}{*}{\vspace{-1.4ex}Method} & \multicolumn{3}{c}{Primary closed-loop objectives} & \multicolumn{2}{c}{Secondary diagnostics} \\ \cmidrule(lr){2-4}\cmidrule(lr){5-6} & CR (\%)$\downarrow$ & OR (\%)$\downarrow$ & AS (m/s)$\uparrow$ & ADE (m)$\downarrow$ & Kin (\%)$\downarrow$ \\
				\midrule
				Pre-trained only          & 2.04$\pm$0.11   & 1.68$\pm$0.10 & 8.36$\pm$0.42 & 1.28$\pm$0.02   & \textbf{0.25$\pm$0.02} \\
				SFT             & 2.01$\pm$0.07   & 1.64$\pm$0.06 & 8.37$\pm$0.36 & \textbf{1.15$\pm$0.015}  & \textbf{0.25$\pm$0.01} \\
				DPO             & 1.97$\pm$0.13   & 1.58$\pm$0.09 & 8.15$\pm$0.39 & 1.33$\pm$0.04   & 0.27$\pm$0.01 \\
				Offline RL      & 2.18$\pm$0.09   & 1.82$\pm$0.14 & \textbf{8.98$\pm$0.68} & 1.37$\pm$0.05   & 0.26$\pm$0.02 \\
				\textsc{SCORP} (Online RL)       & \textbf{1.89$\pm$0.12}   & \textbf{1.36$\pm$0.08} & 8.61$\pm$0.46 & 1.36$\pm$0.04   & 0.32$\pm$0.03 \\
				\bottomrule
		\end{tabular}}
        \vspace{-0.6em}
	\end{table}

    \begin{figure*}[!t]
		\centering
		\includegraphics[width=2.0\columnwidth]{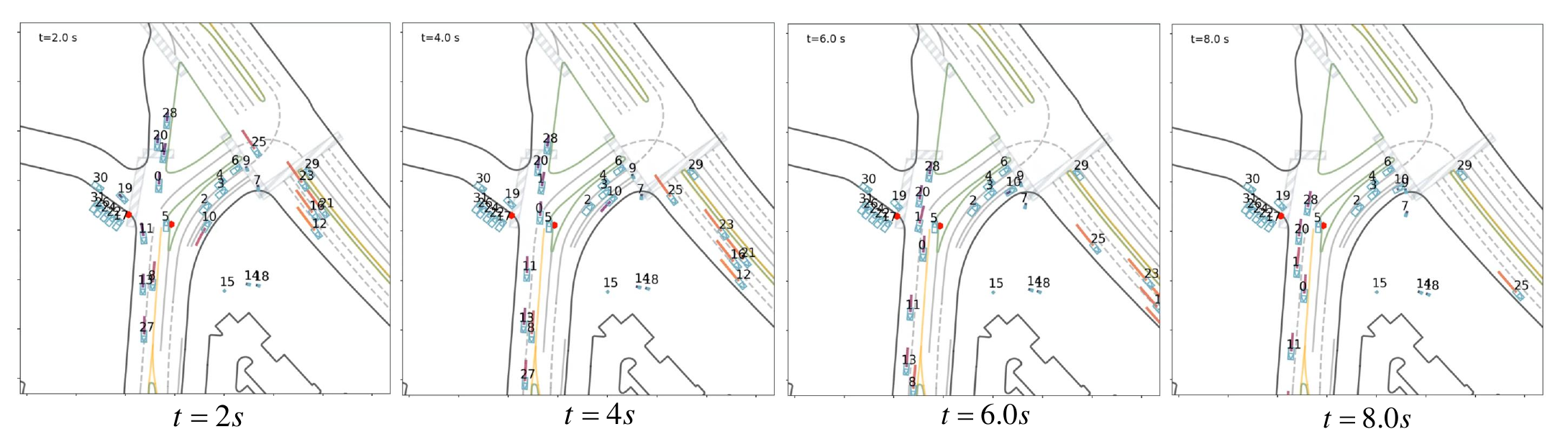}
		\caption{Closed-loop planning visualizations in real traffic scenarios, showing interactive trajectories. We perform closed-loop simulation over an 8-second horizon using a viewpoint centered on Vehicle 1. The 1-second history trajectory is marked behind the vehicles.}
		\label{fig:Visualization}
		\vspace{-0.5em}
	\end{figure*}

    \textbf{Qualitative Visualization.} 
	Fig.~\ref{fig:Visualization} first illustrates the ability of the proposed planner to generate realistic and interactive trajectories in a real intersection scenario involving about 20 vehicles. For clarity, each vehicle is indexed, and the colored traces behind the vehicles show their 1-second motion history. The time-ordered rollout shows that the vehicles can pass through the scene in an orderly and safe manner under the planned trajectories. For example, when traversing a narrow road segment, Vehicles 0, 1, 20, and 28 adjust yielding and acceleration behaviors to maintain safe spacing and complete cooperative passing, indicating that the planner captures multi-vehicle coordination effectively.

    \begin{figure*}[h]
		\centering
		\includegraphics[width=2.0\columnwidth]{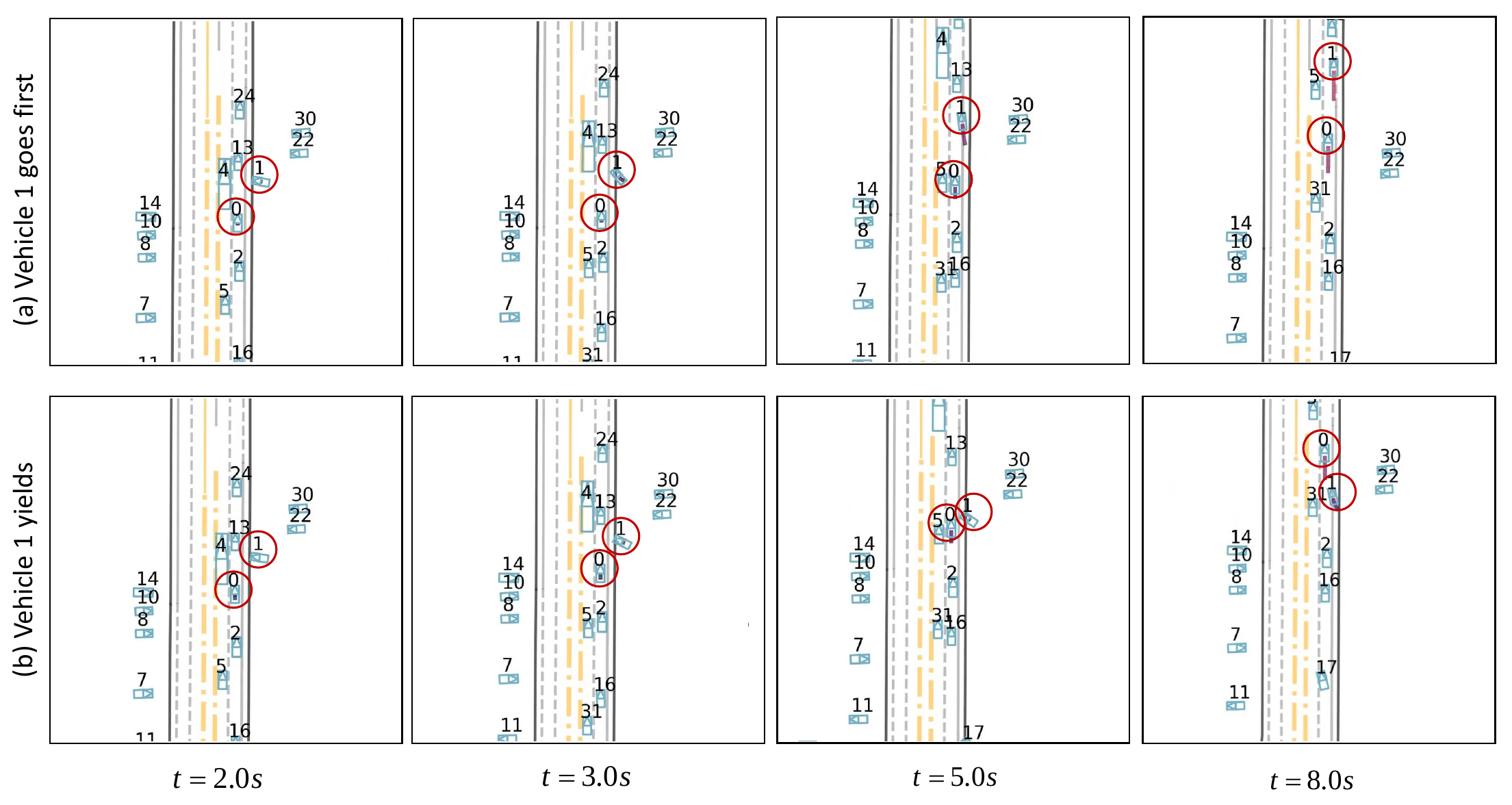}
		\caption{Visualization of diverse interaction behaviors under repeated inference in the same scenario, where Vehicles 0 and 1 exhibit different interaction outcomes: (a) Vehicle 1 goes first; (b) Vehicle 1 yields.}
		\label{fig:multibehavior}
		\vspace{-0.5em}
	\end{figure*}

    We next demonstrate the ability of \textsc{SCORP} to generate diverse cooperative behaviors. Fig. \ref{fig:multibehavior} presents different interaction outcomes obtained from repeated inference in the same scenario. For clarity, Vehicles 0 and 1 are highlighted with red circles, and their historical trajectories are shown. As illustrated in Fig. \ref{fig:multibehavior} (a), Vehicle 1 enters the main road before Vehicle 0 because, in this inference instance, Vehicle 0 is relatively far away, leading Vehicle 1 to take the lead. By contrast, in Fig. \ref{fig:multibehavior} (b), Vehicle 1 yields to Vehicle 0 because, in this instance, Vehicle 0 approaches at a higher speed and from a closer distance, prompting Vehicle 1 to adopt a yielding behavior. These results indicate that \textsc{SCORP} is capable of generating diverse cooperative behaviors, thereby providing diverse samples to support the online reinforcement learning stage for continual improvement of cooperative behavior quality.

    \begin{figure*}[h]
		\centering
		\includegraphics[width=2.0\columnwidth]{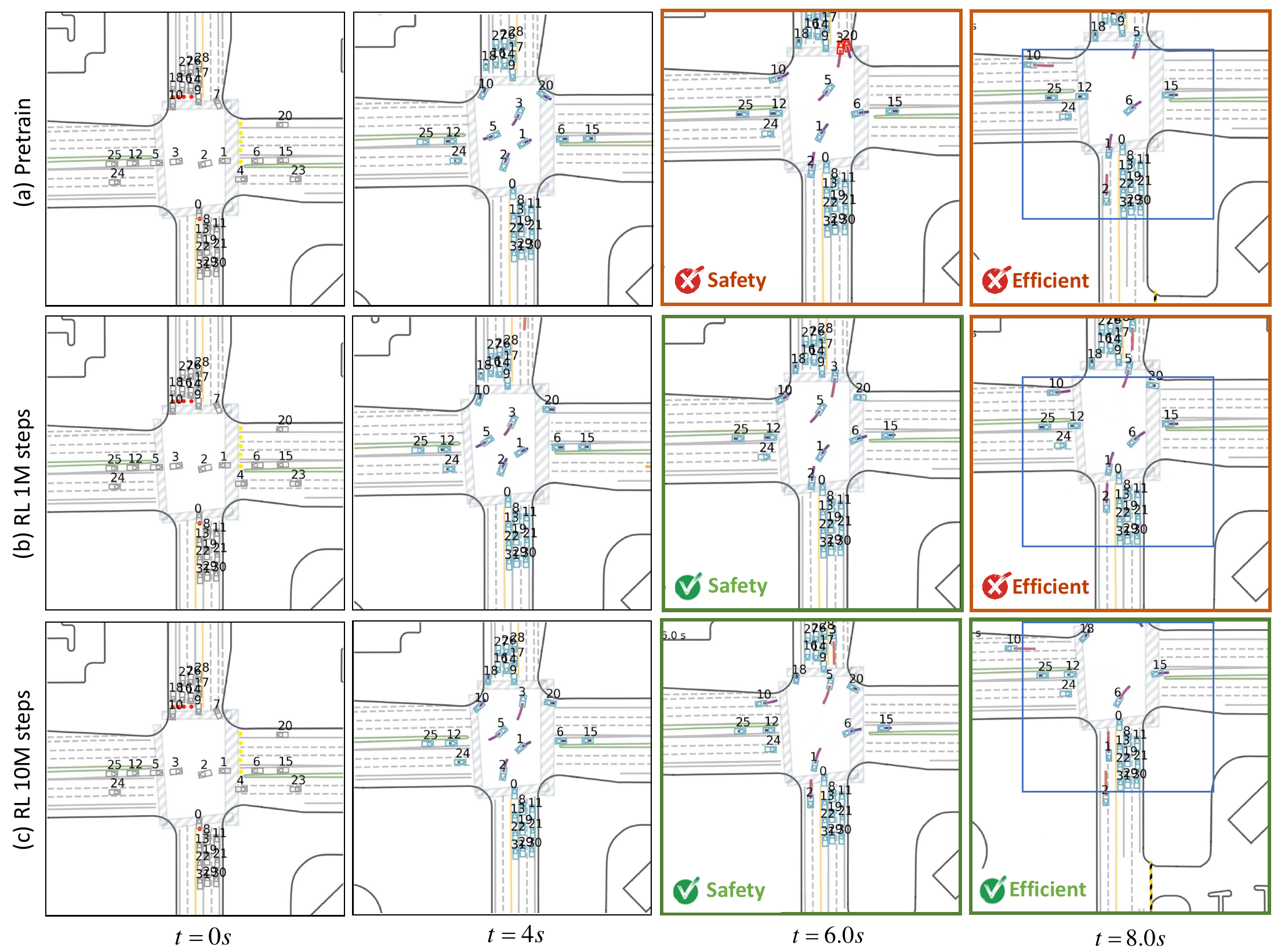}
		\caption{Qualitative comparison between pre-training and RL post-training over an 8-second closed-loop rollout. (a) Pretrained planner. (b) Planner after 1M RL post-training steps. (c) Planner after 10M RL post-training steps. Safety and efficiency steadily improve as post-training progresses.}
		\label{fig:pretrain_posttrain}
		\vspace{-0.5em}
	\end{figure*}

	We further compare pre-training and RL post-training in an out-of-distribution interaction case. As shown in Fig.~\ref{fig:pretrain_posttrain} (a), the pre-trained model generates a left-turn trajectory for Vehicle 3 but fails to adequately resolve the interaction conflict between Vehicle 3 and Vehicle 20, resulting in a collision at $t=6.0$ s. In Fig.~\ref{fig:pretrain_posttrain}(b), after 1M RL steps, the model learns a more conservative yielding strategy in the challenging left-turn scenario, avoiding collisions at the cost of reduced traffic efficiency. In Fig.~\ref{fig:pretrain_posttrain} (c), after 10M RL steps, the model preserves safety while recovering more flexible and decisive interaction behaviors, and generates more efficient trajectories for moving vehicles such as Vehicles 0, 1, 2, 6, 10, and 18. These qualitative cases show that reward-driven RL post-training progressively reshapes the interaction strategy of the planner, improving the balance between safety and efficiency in closed-loop multi-agent driving.

	\subsection{Ablations}\label{sec:ablation}
	
	Unless otherwise stated, ablations use the same training pipeline and evaluation protocol as Section~\ref{sec:results}.

    \textbf{Impact of AdaLN-Zero Module.} AdaLN-Zero consistently improves conditional generation quality and closed-loop safety. 
    As shown in Table~\ref{tab:adaln_ablation}, introducing AdaLN-Zero reduces the collision rate from 2.11\% to 2.04\% and, more notably, lowers the off-road rate from 2.05\% to 1.68\%, indicating substantially improved scene consistency and boundary adherence. This improvement arises because condition-driven modulation reshapes denoising features and strengthens awareness of road geometry, enabling the model to exploit scene cues more effectively than cross-attention alone and to generate trajectories that better comply with scene constraints. The resulting improvement in boundary compliance further translates into stronger closed-loop stability.

\begin{figure}[h]
		\centering
		\vspace{-5pt}
		\subfigure[without AdaLN-Zero module\label{fig:without_adaln}]{
			\includegraphics[width=0.47\columnwidth]{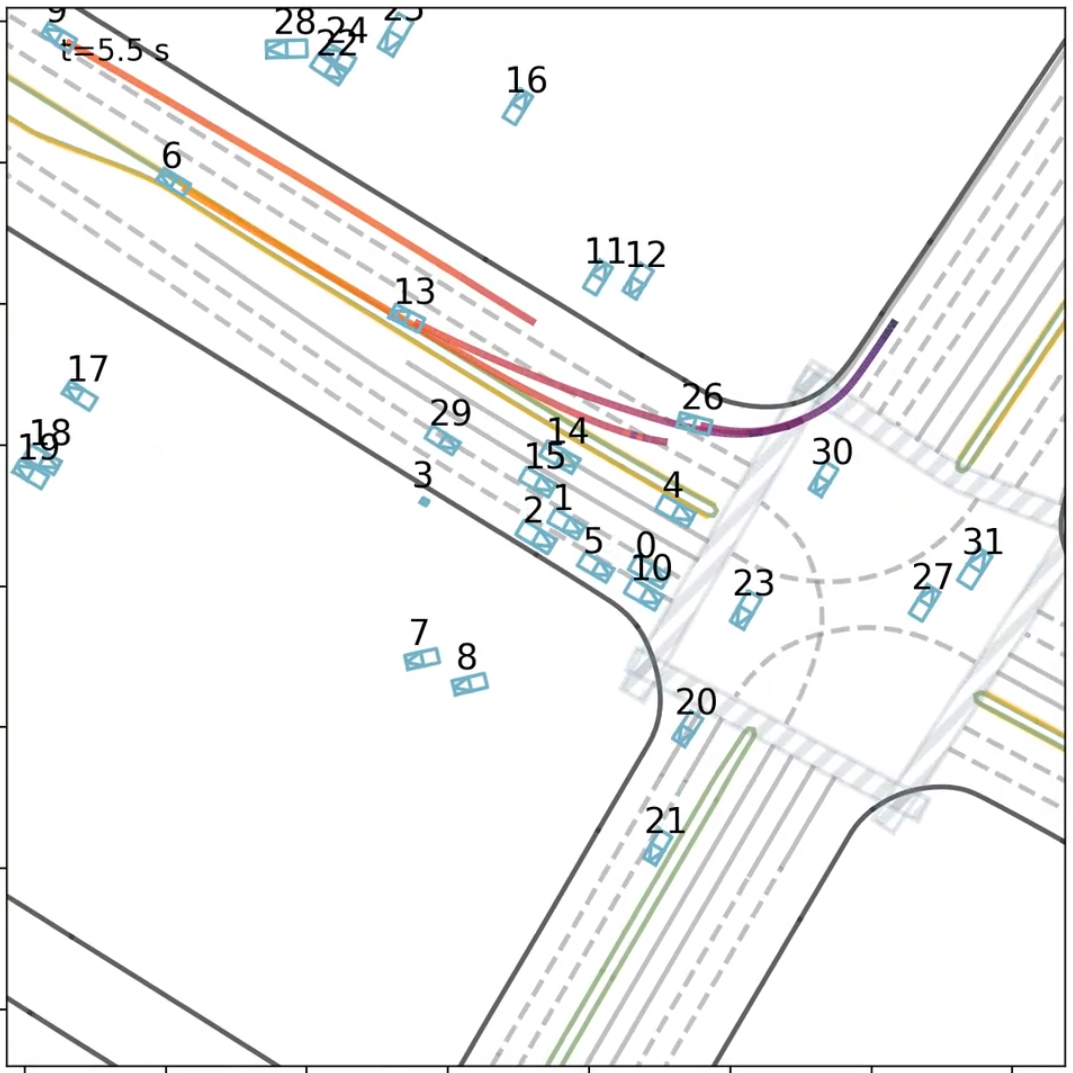}
		}\hfill
		\subfigure[with AdaLN-Zero module\label{fig:with_adaln}]{
			\includegraphics[width=0.47\columnwidth]{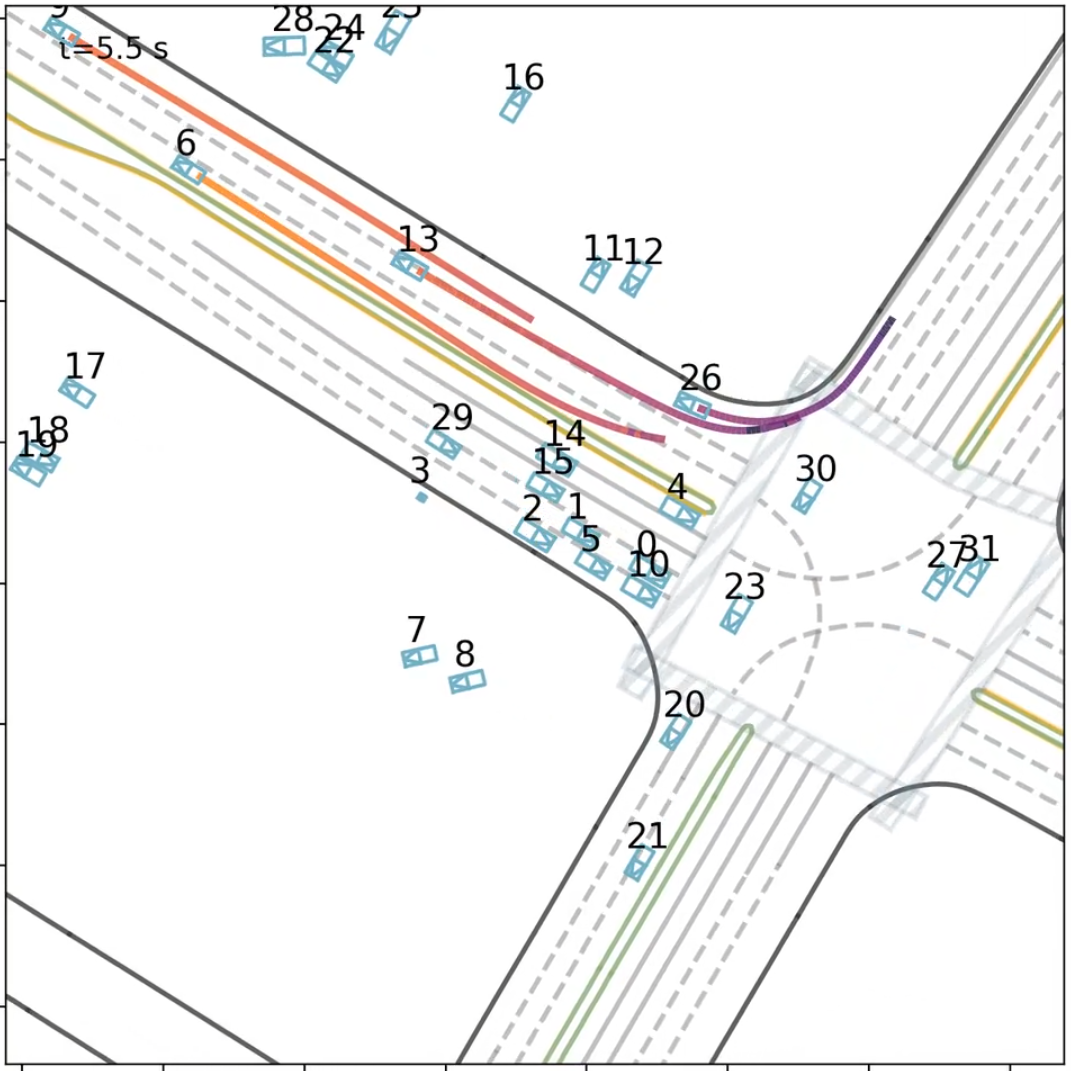}
		}
		\caption{Qualitative comparison with and without the AdaLN-Zero module}
		\label{fig:adaln}
	\end{figure}

	Fig.~\ref{fig:adaln} provides a qualitative comparison on sharp right-turn cases. Without AdaLN-Zero, Vehicles 6 and 13 drift outward and hit the median boundary. With AdaLN-Zero, turn trajectories are more compact and maintain safer boundary margins.
	
	\begin{table}[!t]
		\caption{Ablation on AdaLN-Zero module}\label{tab:adaln_ablation}
		\vspace{-0.5em}
		\centering\footnotesize
		\setlength{\tabcolsep}{3pt}
		\resizebox{0.99\columnwidth}{!}{%
			\begin{tabular}{lccccc}
				\toprule
				Setting &   CR (\%)$\downarrow$ & OR (\%)$\downarrow$ & AS (m/s)$\uparrow$ & ADE (m)$\downarrow$ & Kin (\%)$\downarrow$ \\
				\midrule
				w/o AdaLN-Zero         & 2.11  & 2.05 & \textbf{8.40} & 1.30   & 0.26 \\
				with AdaLN-Zero             & \textbf{2.04}   & \textbf{1.68}  & 8.36 & 1.28   & 0.25 \\
				\bottomrule
		\end{tabular}}
	\end{table}

	\textbf{Impact of the Variance-gated Mechanism.} 
	Table~\ref{tab:vg_grpo} analyzes the effect of our \emph{variance-gated} mechanism via ablations over several settings, including a non-gated baseline and multiple threshold combinations. Without variance gating, post-training collapses at $\sim$0.5M steps, evidenced by persistently large gradient-norm fluctuations and a sudden spike in the KL term. The last stable checkpoint also exhibits degraded performance across metrics.
	We then vary the gating thresholds. With $\text{std}_1/\text{std}_2=0.00/0.06$, collapse is delayed to $\sim$2.0M steps and key closed-loop safety and efficiency metrics improve slightly, indicating that $\text{std}_2$ contributes to training stability. In contrast, $0.03/0.06$ and $0.03/0.09$ prevent collapse completely, suggesting that $\text{std}_1$ is critical for stability by mitigating advantage degeneration when within-group samples are nearly identical. Among stable settings, $0.03/0.06$ achieves the best performance. A larger $\text{std}_2$ appears overly conservative and weakens the effective learning signal, suggesting that moderate thresholds strike a better balance between stability and performance gains.

		\begin{table}[!t]
			\caption{Ablation of the variance-gated mechanism (performance is evaluated at the last checkpoint before collapse)}\label{tab:vg_grpo}
			\vspace{-0.5em}
			\centering\footnotesize
		\setlength{\tabcolsep}{3pt}
		\resizebox{0.99\columnwidth}{!}{%
			\begin{tabular}{lcccccc}
				\toprule
				Gating $\text{std}_1$/$\text{std}_2$ & Collapse step & CR$[\%]$$\downarrow$ & OR$[\%]$$\downarrow$ & AS$[m/s]$$\uparrow$ & ADE$[m]$$\downarrow$ & Kin$[\%]$$\downarrow$ \\
				\midrule
				w/o  gating & $\approx$ 0.5M & 2.15 & 2.03 & 8.05 & 1.74 & 0.40 \\
				0.03/0.06 & -- & \textbf{1.89} & \textbf{1.36} & \textbf{8.61} & 1.36 & 0.32 \\
				0.00/0.06 & $\approx$ 2.0M & 2.01 & 1.57 & 8.42 & 1.42 & 0.30 \\
				0.03/0.09 & -- & 1.96 & 1.50 & 8.49 & \textbf{1.30} & \textbf{0.29} \\
				\bottomrule
		\end{tabular}}
		\vspace{-0.9em}
		\end{table}

		\textbf{Impact of Post-training Data Distribution.}
		Table~\ref{tab:data_dist_ablation} analyzes how the composition of post-training scenes affects final performance. Training only on the high-score subset yields limited gains, because this subset contains many relatively simple scenes and thus provides weak optimization signals due to insufficient diversity among sampled trajectories. By contrast, training only on the low-score subset substantially increases the collision and off-road rates, while yielding only a slight increase in average speed, indicating clear degradation in overall performance. This suggests that when post-training data are overly concentrated on hard failure cases, policy updates may be dominated by more explicit local efficiency-oriented corrective objectives and fail to establish a stable global optimization direction.
		The best overall result is obtained on the full dataset, which mixes high-score, low-score, and regular scenes. This observation indicates that a balanced post-training data distribution is important for stable reinforcement post-training: it maintains broader policy coverage and more stable sampling behavior, and thus provides more informative gradient signals for effective optimization and better generalization.
		
		\begin{table}[!t]
			\caption{Ablation on post-training data distribution}\label{tab:data_dist_ablation}
			\vspace{-0.5em}
			\centering\footnotesize
			\setlength{\tabcolsep}{4pt}
			\resizebox{0.98\columnwidth}{!}{%
				\begin{tabular}{lccccc}
					\toprule
					Dataset type & CR (\%)$\downarrow$ & OR (\%)$\downarrow$ & AS (m/s)$\uparrow$ & ADE (m)$\downarrow$ & Kin (\%)$\downarrow$ \\
					\midrule
					High-score & 1.95 & \textbf{1.34} & 8.53 & \textbf{1.30} & \textbf{0.31} \\
					Low-score & 2.18 & 1.99 & 8.49 & 1.40 & 0.37 \\
					Full & \textbf{1.89} & 1.36 & \textbf{8.61} & 1.36 & 0.32 \\
					\bottomrule
			\end{tabular}}
			\vspace{-0.5em}
		\end{table}

		\section{Conclusion}\label{sec:conclusion}
	
	In this paper, we introduced \textsc{SCORP}, a cooperative multi-agent planner that couples condition-enhanced diffusion pre-training with stable online RL post-training in reactive closed-loop environments. For pre-training, \textsc{SCORP} strengthens scene consistency and road adherence by combining inter-agent self-attention with dual-path scene conditioning. For post-training, \textsc{SCORP} achieves stable online learning by combining dense, well-shaped rewards with our proposed VG-GRPO, mitigating advantage collapse and gradient instability while strengthening closed-loop cooperative behaviors. Extensive experiments demonstrate superior closed-loop planning performance on key \textit{safety} and \textit{efficiency} metrics, outperforming state-of-the-art baselines on WOMD and across alternative post-training paradigms.

	\bibliographystyle{IEEEtran}
	\bibliography{se05_refs_normalized}
	
	\end{document}